\pdfoutput=1
\documentclass[11pt]{article}

\usepackage{EMNLP2022}

\usepackage{times}
\usepackage{latexsym}
\usepackage{booktabs}
\usepackage{comment}
\usepackage{graphicx}
\usepackage{subcaption}
\usepackage[T1]{fontenc}
\usepackage[utf8]{inputenc}

\usepackage{microtype}
\newcommand{\sara}[1]{\textcolor{blue}{sara: #1}}
\newcommand{\julia}[1]{\textcolor{magenta}{julia: #1}}

\newcommand{\wikiann}[0]{WikiAnn }

\usepackage{inconsolata}

\usepackage{enumitem}

\title{Intriguing Properties of Compression on Multilingual Models}
% ...

\author{Kelechi Ogueji\thanks{Work was done while at Google Research.}\\
  University of Waterloo \\
  \texttt{kjogueji@uwaterloo.ca} \\
  \\\And
  Orevaoghene Ahia \\
  University of Washington \\
  \texttt{oahia@cs.washington.edu} \\
  \\\And
  Gbemileke Onilude  \\
  Cohere For AI Community \\
  \texttt{lekeonilude@gmail.com} \\
  \\\AND
  Sebastian Gehrmann\\
  Google Research \\
  \texttt{gehrmann@google.com} \\
   \\\And
  Sara Hooker \\
  Cohere For AI \\
  \texttt{sarahooker@cohere.com} \\
   \\\And
 Julia Kreutzer \\
  Google Research \\
  \texttt{jkreutzer@google.com} \\
  }

\begin{document}
\maketitle
\begin{abstract}
% Given the rapidly growing size of pre-trained language models, there has been an increased focus on compressing these models.
% %However, much of this work has been geared towards monolingual models.
% Understanding the impact that compression has on model performance is especially important in multilingual settings, where prior work indicates that low-resource languages may be more negatively impacted.
Multilingual models are often particularly dependent on scaling to generalize to a growing number of languages. Compression techniques are widely relied upon to reconcile the growth in model size with real world resource constraints, but compression can have a disparate effect on model performance for low-resource languages. It is thus crucial to understand the trade-offs between scale, multilingualism, and compression. In this work, we propose an experimental framework to characterize the impact of sparsifying multilingual pre-trained language models during fine-tuning.
Applying this framework to mBERT named entity recognition models  across 40 languages, we find that compression confers several intriguing and previously unknown generalization properties. In contrast to prior findings, we find that compression may improve model robustness over dense models. We additionally observe that under certain sparsification regimes compression may aid, rather than disproportionately impact the performance of low-resource languages.
% Our findings have several real-world implications for machine learning practitioners and provides interesting directions for future works on sparsifying multilingual models.
\end{abstract}

\section{Introduction}

Scaling language models benefits multilingual settings, since it is difficult to maintain performance across a growing number of languages at a constant model size, a property also called the ``curse of multilinguality''~\citep{NEURIPS2019_c04c19c2,conneau-etal-2020-unsupervised,Artetxe2019MassivelyMS}.
However, the extent of growth in language model (LM) size ~\citep{Radford2019LanguageMA,Brown2020LanguageMA,Zhang2022OPTOP, Chowdhery2022PaLMSL} has made deployment to resource-constrained environments much more challenging~\citep{warden2019tinyml,Samala_2018,treviso2022}. 
To benefit from the performance gains conferred by scale, efficiency techniques that reduce model size while maintaining comparable aggregate performance are widely used, such as quantization~\citep{shen2020q}, compression~\citep{michel2019sixteen,lagunas-etal-2021-block} and distillation~\citep{tsai-etal-2019-small,sanh2019distilbert,pu-etal-2021-learning}.

While most compression techniques have minimal impact on aggregate performance numbers ~\citep{2019arXiv190209574G,Li2020LearningLT,Hou2020DynaBERTDB,Chen2021EarlyBERTEB,Bai2020BinaryBERTPT,tessera2021gradients}, the impact on individual sub-populations in the data, such as low-resource languages,  can be far more severe~\citep{2019shooker,2020hooker,ahia-etal-2021-low-resource}. Disparities in resource availability become more apparent at larger scale, both in terms of data and deployment resource availability. This makes compression all the more necessary, but also motivates a thorough consideration of the subsequent impact of compression on generalization.

In this work, we develop an experimental framework to investigate \textit{the impact of compression during fine-tuning of pre-trained multilingual models} which we apply to Named Entity Recognition (NER) across 40 languages of the \wikiann  benchmark~\citep{pan-etal-2017-cross}. We study the impact of compression on groups of languages across multiple dimensions---resourcedness, script, and language family---and evaluate the sensitivity of models to input perturbations along these groupings. % by proposing and crafting evaluation benchmarks which ensure coverage across all languages considered.

%Evaluating the quality of the models on original and perturbed test sets leads to several surprising findings:
This leads us to discover the following \emph{intriguing properties}: (1) Lower-performing languages disproportionately suffer under extreme levels of sparsity, as pruning amplifies disparities. However, low-resource languages present an intriguing \textit{flip-flop} moment, where their performance may benefit from medium regimes of sparsity.
(2) We find that dense models overfit to typical test cases, achieving a close-to-0 F1 score on slightly perturbed inputs, while compression can recover close to the original test performance. Our results stand in contrast to previous work that find that sparsity erodes robustness, suggesting more work is needed to understand the dynamics between compression and robustness. 
(3) The choice to prune model embeddings can completely negate the two benefits described in the previous observations, showing the importance of comparing the two cases in future analyses.

\begin{comment}

\begin{figure}[htbp]
\includegraphics[width=\columnwidth]{plots/f1_score_mean_vs_sparsity_without_embeddings.pdf}
\caption{F1 score across levels of sparsity and grouped by resource level.}
\end{figure}

% With embeddings

\begin{figure}[htbp]
\includegraphics[width=\columnwidth]{plots/f1_score_mean_vs_sparsity_with_embeddings.pdf}
\caption{F1 score across levels of sparsity and grouped by resource level.}
\end{figure}

\begin{figure}[htbp]
\includegraphics[width=\columnwidth]{plots/f1_score_diff_mean_vs_sparsity_without_embeddings.pdf}
\caption{F1 score difference between various levels of sparsity and the zero-sparsity model grouped by resource level.}
\end{figure}

\begin{figure}[htbp]
\includegraphics[width=\columnwidth]{plots/f1_score_mean_diff_vs_sparsity_with_embeddings.pdf}
\caption{F1 score difference between various levels of sparsity and the zero-sparsity model grouped by resource level.}
\end{figure}

\end{comment}

\section{Related Work}

The ``curse-of-multilinguality'' creates a tradeoff between number of languages and size of a model~\citep{conneau-etal-2020-unsupervised}. However, training smaller models supporting fewer languages may not always be feasible~\citep{abdaoui-etal-2020-load}. 
Compressing large models has been shown to combat the curse, either by compressing the pretrained model~\citep{tsai-etal-2019-small,sanh2019distilbert}, or by compressing during finetuning, as in our case. 
While many studies investigate the impact of pruning on aggregate metrics in monolingual pre-trained LMs~\cite{NEURIPS2020_eae15aab,Goyal2020PoWERBERTAB,gordon-etal-2020-compressing, budhraja-etal-2020-weak,Sajjad2020PoorMB,lagunas-etal-2021-block,xu-etal-2021-beyond,Du2021WhatDC,ganesh-etal-2021-compressing}, fewer works focus on multilingual settings \cite{mukherjee-hassan-awadallah-2020-xtremedistil,ansell-etal-2022-composable}. 
Yet, prior analyses find a disparate effect of removing attention heads or model layers on languages and language families distant from the training data in NER~\citep{ma-etal-2021-contributions, Budhraja2021OnTP}, demonstrating the importance of looking into subpopulations as we do in this study. Concurrent work \cite{mohammadshahi2022compressed} explores compression on multilingual machine translation models, and find that while under-represented languages suffer extreme performance drop, some medium-resource language benefit from compression. However, unlike our proposed framework, they perform this study without additional fine-tuning after compression.

Studies that compare the robustness of compressed and dense models further find that compression may lead to erosion of performance on ``challenging'' samples and poor generalization~\citep{ahia-etal-2021-low-resource,Du2021WhatDC,xu-etal-2021-beyond}, a finding that we expand on and connect to language resourcedness. The technique we use to study robustness expands on studies that perturb training~\citep{Yaseen2021DataAF,dai-adel-2020-analysis} or evaluation data~\citep{dhole2021nl} in NER by introducing perturbations specific to languages, language families, and scripts.

\section{Methodology}
%To address this, 
% We propose an experimental framework to measure the effects of sparsity when fine-tuning pre-trained multilingual LMs.
% Focusing on NER across 40 languages, we investigate the impact of sparsity on a per-language, family and script basis, all within monolingual and multilingual settings.
% Furthermore, we investigate the robustness of sparse multilingual LMs to perturbations in the dataset.
% We select NER for two major reasons: Firstly, we are able to obtain data for a multitude of languages, and secondly we are able to easily perform perturbations, which we will discuss later in Section~\ref{sec:finetuning}.\julia{this paragraph might be redundant enough to cut it for space} \sg{agree}

% We next describe our experimental framework to investigate the impact of sparsity when introduced during fine-tuning of pre-trained multilingual LMs. % on a per-language, family, and script basis within monolingual and multilingual settings.
% \julia{could be cut for space if needed}

\paragraph{Data}\label{sec:data} We conduct our experiments on \wikiann~\citep{pan-etal-2017-cross}, a multilingual NER dataset. \wikiann was sourced from Wikipedia articles and automatically annotated with LOC (location), PER (person), and ORG (organisation) labels in the IOB2 format \citep{ramshaw-marcus-1995-text}. It is considered a ``silver standard'' due to its automatic entity labels and noise~\citep{lignos-etal-2022-toward}, but with its 176 languages it covers the most languages of any NER dataset. We focus our experiments on the 40 languages from the XTREME benchmark~\citep{hu2020xtreme},
with train-test splits defined by \citet{rahimi-etal-2019-massively}. These training sets were built with stratified sampling to create a balance across entity types~\citep{lignos-etal-2022-toward}, and are thus a subset of the total available data from the original WikiAnn. Table~\ref{tab:data} lists language codes in  ISO 639-1 and their available training data for fine-tuning.
%\footnote{The percentage of tokens for a language during mBERT pre-training and the number of sentences for fine-tuning for \wikiann are correlated (Kendall's $\tau=0.64$)---those languages that have higher representation during pre-training tend to also have larger amounts of fine-tuning data.} % \oreva{Should we make this a footnote ?}\julia{good idea}
%\julia{don't think it's essential after all, it's quite visible from the table}
%\sara{I personally do not like footnotes in paper, I find it disrupts the flow of the reader -- I think anything we want in a footnote deserves to be in the main text} \sg{strongly disagree with this. Footnotes help not interrupting the main flow - \url{https://www.cs.jhu.edu/~jason/advice/in-defense-of-footnotes.html} }\julia{on Sebastian's side here :) Oreva also suggested this initially, so I think the majority is in favor of the footnote.} \sara{lol, I love how sebastians links to a defense of the footnote. ok I will give up this battle -- but I do continue to disagree.}

% \sara{I moved the perturbations section around to gain some space}%\julia{shouldn't this be part of the Evaluation subsection? or it's own subsection}  
\paragraph{Perturbations} We test the robustness of compressed models by perturbing named entities in the test set. Previous work \citep{Du2021WhatDC} show that sparse pretrained language models are less robust than their dense equivalents when evaluated on adversarial test sets, even when they perform similarly on in-distribution test sets. We adopt a data perturbation technique from \citet{dai-adel-2020-analysis} called \textit{entity mention replacement}; an entity is randomly swapped with another entity of the same type (example sentences shown in App. \ref{app:examples}). 
We first perturb entities within same language for all the languages in our dataset (\texttt{in-language}). Secondly, we propose a new benchmark appropriate for testing the \emph{cross-lingual robustness} of multilingual models on our downstream task. We perturb entities across different languages that share common linguistic properties. In particular, we group languages by family and script and perturb entities across languages within the same group (\texttt{in-script}, \texttt{in-family}).

%  \julia{what will the evaluation on these perturbations be able to tell us about our models?} \sg{Can we further state that adding completely different entities in different scripts may just be too easy since it is essentially a shortcut for the model to just tag everything that is different script? hence, our approach is better?}
%  \sara{I don't think it is clear it is easier given that the model cannot train on the perturbed data. As far as I understand we are doing eval on perturbed only.}
%  \julia{I think Sebastian means that the model might have already learned on the usual training data that tokens from a different script are usually entities. This is probably the case for non-Latin-script languages, where Latin script might be used for foreign entities.}
% Additionally, we evaluate the impact of sparsity on the robustness of the finetuned multilingual model by evaluating sensitivity to entity perturbations. To do this, we propose a new evaluation benchmark appropriate for a multilingual setting which is  described below. %Section~\ref{sec:evaluation}.
%Although WikiANN contains 176 languages, we select 40 of them from different language families and scripts.
\begin{table}[t]
    \centering
        \resizebox{\columnwidth}{!}{%
    \begin{tabular}{l|lc}
    \toprule
    \# Sent. & Languages & Pretr. Token \%\\
    \midrule
         100  & \texttt{\underline{jv},my,\underline{yo}}&   0.05\\
1000  & \texttt{kk,\underline{sw},\underline{te}}&  0.19 \\
5000  & 	\texttt{\underline{af},\underline{hi},mr} &  0.21\\
10000  & \texttt{\underline{bn},\underline{eu},ka,ml,tl}&  0.23\\
15000  &\texttt{et,ta}&  0.31\\
20000  &\texttt{ar,bg,de,el,\underline{en},es,fa,fi} & 2.93\\
& \texttt{fr,he,hu,id,it,ja,ko,ms} & \\
& \texttt{nl,pt,ru,th,tr,ur,vi,\underline{zh}} & \\
    \bottomrule
    \end{tabular}%
    }
    \caption{Data sizes and languages for \wikiann and average representation for mBERT pre-training. The underlined languages are used for a comparison with monolingual fine-tuning.}
    \label{tab:data}
\end{table}

\paragraph{Model}\label{sec:model} We use the cased multilingual BERT (mBERT)~\citep{devlin-etal-2019-bert} for all our experiments because it is one of the most widely used and studied multilingual LMs~\citep[e.g.,][]{pires-etal-2019-multilingual,ronnqvist-etal-2019-multilingual,wu-dredze-2019-beto,wang-etal-2020-extending,chi-etal-2020-finding}%chau-etal-2020-parsing}%,dufter-schutze-2020-identifying,tanti-etal-2021-language,pan-etal-2021-multilingual,muller-etal-2021-unseen,rajaee-pilehvar-2022-isotropy}
.\footnote{While XLM-R~\citep{conneau-etal-2020-unsupervised} and others may perform better, the availability of fine-grained mBERT results through XTREME~\citep{hu2020xtreme} allowed us to start from parameters that replicate the prior results.} mBERT is trained on Wikipedia data from 104 languages, has approximately 177 million parameters, and a vocabulary size of around 120,000. %To account for resource-differences between languages,
%mBERT applies an exponentially smoothed weighting of the data which leads to up-sampling data from lower-resource languages and down-sampling those from higher-resource languages.
We finetune mBERT by appending a linear classification layer to the model and updating all its parameters. Full hyperparameters are listed in App.~\ref{app:ft_hyper}.
%\julia{add section in the appendix and refer to it here} 
%\sg{Mention how many seeds / replications}

%\paragraph{Monolingual vs. Multilingual}
We evaluate the impact of sparsity in two settings: 1) In the \texttt{monolingual} setting we fine-tune on individual languages. For this setting, we select 10 languages with different available data size, language family and scripts (see Table~\ref{tab:data}). 2) In the \texttt{multilingual} setting we jointly fine-tune on all languages. We train all models with three random seeds, and evaluate F1 using \texttt{seqeval}~\citep{seqeval} on the individual languages' evaluation data. We report mean results across runs after computing the micro-average F1 scores across entity classes.
%the mean results across runs.
%We report mean results over these repeated runs in all following analyses.

%In both settings, we perform evaluation on the individual languages' evaluation data.

%in two settings: monolingual and multilingual.
%In the monolingual setting, we fine-tune mBERT on in-language training data only.
%However, 
%In the multilingual setting, we fine-tune mBERT on a multilingual data obtained from a concatenation of all languages' training data.

\begin{figure}[t]
    \centering
    \includegraphics[width=\columnwidth]{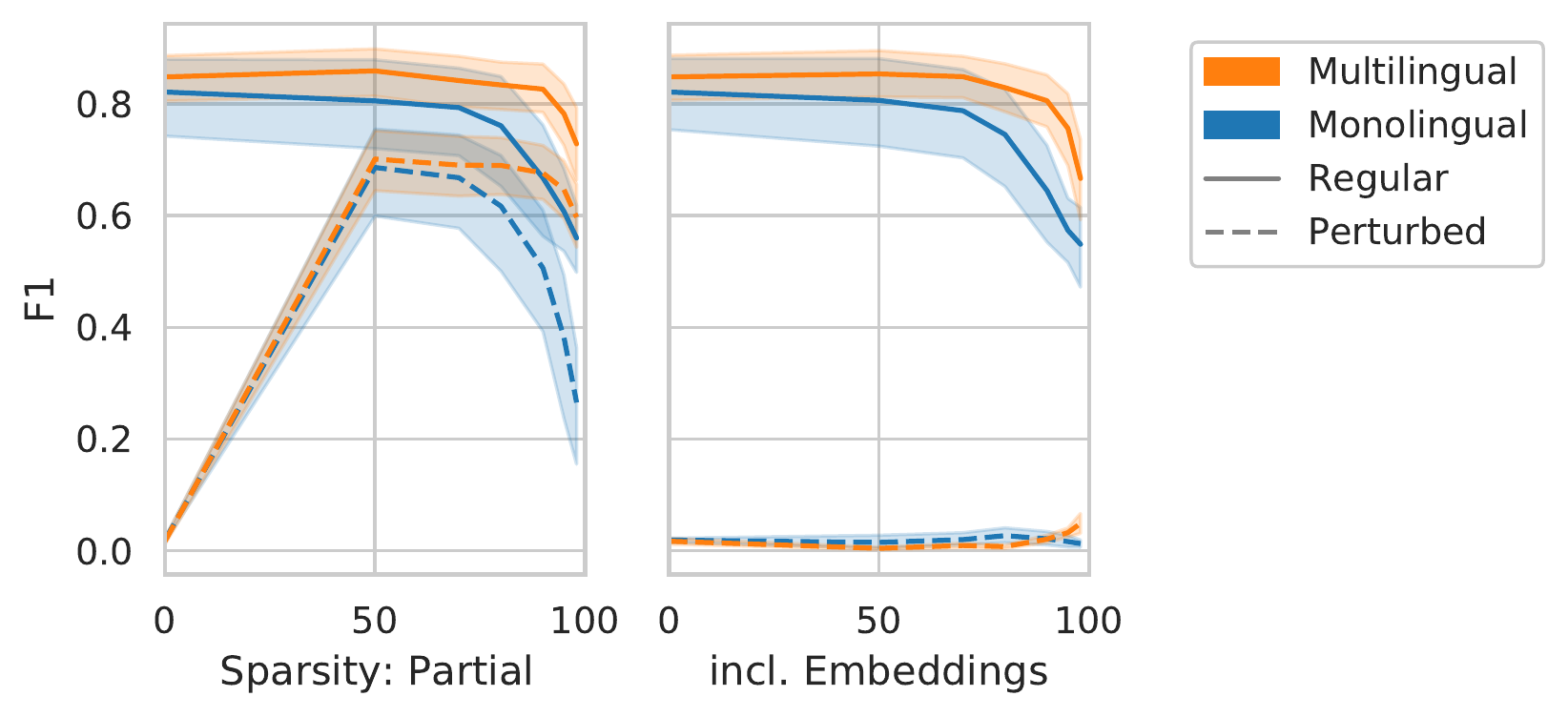}
    \caption{\textbf{Monolingual vs Multilingual:} F1 for monolingual and multilingual fine-tuning under regular and perturbed test conditions (\texttt{in-language}), averaged across languages (shaded areas: standard deviation).}
    \label{fig:multilingual}
\end{figure}

\paragraph{Pruning}
We induce sparsity by applying Iterative Magnitude Pruning (IMP)~\citep{Han2015,han2016deep} during fine-tuning. IMP iteratively removes weights that are below a certain threshold until a desired target sparsity is reached. IMP is widely used and competitive with far more compute intensive approaches \citep{gale2019,gordon-etal-2020-compressing,mengnan-forget,ganesh-etal-2021-compressing}, while allowing us to sparsify to an exact level.
% \julia{need support for that: in NLP? cite papers} \sg{This is the third time we mention "we use it because it is popular". Do we have an explanation for why it is the correct one to use? I really dislike papers that just blindly follow what others do but don't argue for their choices. Even wrote a paper about it (https://arxiv.org/abs/2202.06935)}
% \sara{it is competitive with state of art (on par or better for most ranges of sparsity until 90, and comparable for sparsity ranges above that despite being one of the simplest methods to implement. It only requires simple thresholding and no auxiliary models. This is why it is very widely used -- probably the most common method by far. It is also widely used because you can pre-fix the level of sparisty you want, unlike in regularization based methods where you essentially can't precisely control the end sparsity -- this is important for this particular experimental set-up. Let's take a go at folding in some of this reasoning -- Kelechi/Oreva can you give it a try in your edits tomorrow.}
%\julia{totally agreed}
%We apply IMP during fine-tuning by gradually pruning model parameters until a target sparsity.
We compare two pruning strategies: 1) \texttt{partial} where we prune all dense layers except for embedding layers, 2) \texttt{incl. embeddings} where we prune all dense weights including embedding layers. Embeddings make up more than half (91M) of the 177M parameters in mBERT, while dense weights make up the rest. Hence, pruning embeddings allows us to significantly reduce the number of mBERT parameters.
We consider five sparsity levels: 50\%, 70\%, 80\%, 90\%, 95\% and 98\%, corresponding to the percentage of weights pruned (hyperparameters in App.~\ref{app:prune_hyper}. Preliminary experiments were conducted with lower sparsity levels (10\%-40\%) and yielded similar findings to those at moderate sparsity levels (50\%-70\%), motivating the sparsity intervals chosen. The chosen sparsity levels also align with general best practice in sparsity evaluation as presented in previous works. Moderate to high sparsity levels (50\%+) are necessary for efficiency gains in the real-world and are usually studied in literature \cite{gale2019,ahia-etal-2021-low-resource,ganesh-etal-2021-compressing}.

% This allows us to significantly reduce the number of parameters of mBERT as opposed to pruning only dense layers.
%\sara{naming is a little clunky, but we should differentiate between these settings.}
%\julia{how about "partial" and "incl. embeddings"? this is what we use in the plots.}
%Pruning is done in two settings: in the first setting, we prune only dense weights and avoid the embedding layers, while in the second setting, we prune both dense weights and the embedding layers.
% \julia{need to create a section for it and refer to it}
%\julia{TODO Kelechi: discuss whether embeddings should be included or not, according to literature and how much it affects model size}

%\paragraph{Evaluation} 
% \sg{For space-saving, can we merge this with the above?}
%\julia{done}

\begin{comment}
\begin{figure*}
    \centering
    \includegraphics[width=\textwidth]{plots/abs_f1_ftsize_relplot.pdf}
    \caption{Mean absolute difference in F1 compared to the dense model. Results are averaged for languages grouped according to fine-tuning size. The shaded areas represent the standard deviation. Results in the first row are computed on the regular test sets, those in the second row on the in-language perturbed test sets. The models in the first column do not have pruned embeddings, in the second column the embeddings are pruned as well.}
    \label{fig:abs_pruning}
\end{figure*}
\end{comment}

\begin{figure}
    \centering
    \includegraphics[width=\columnwidth]{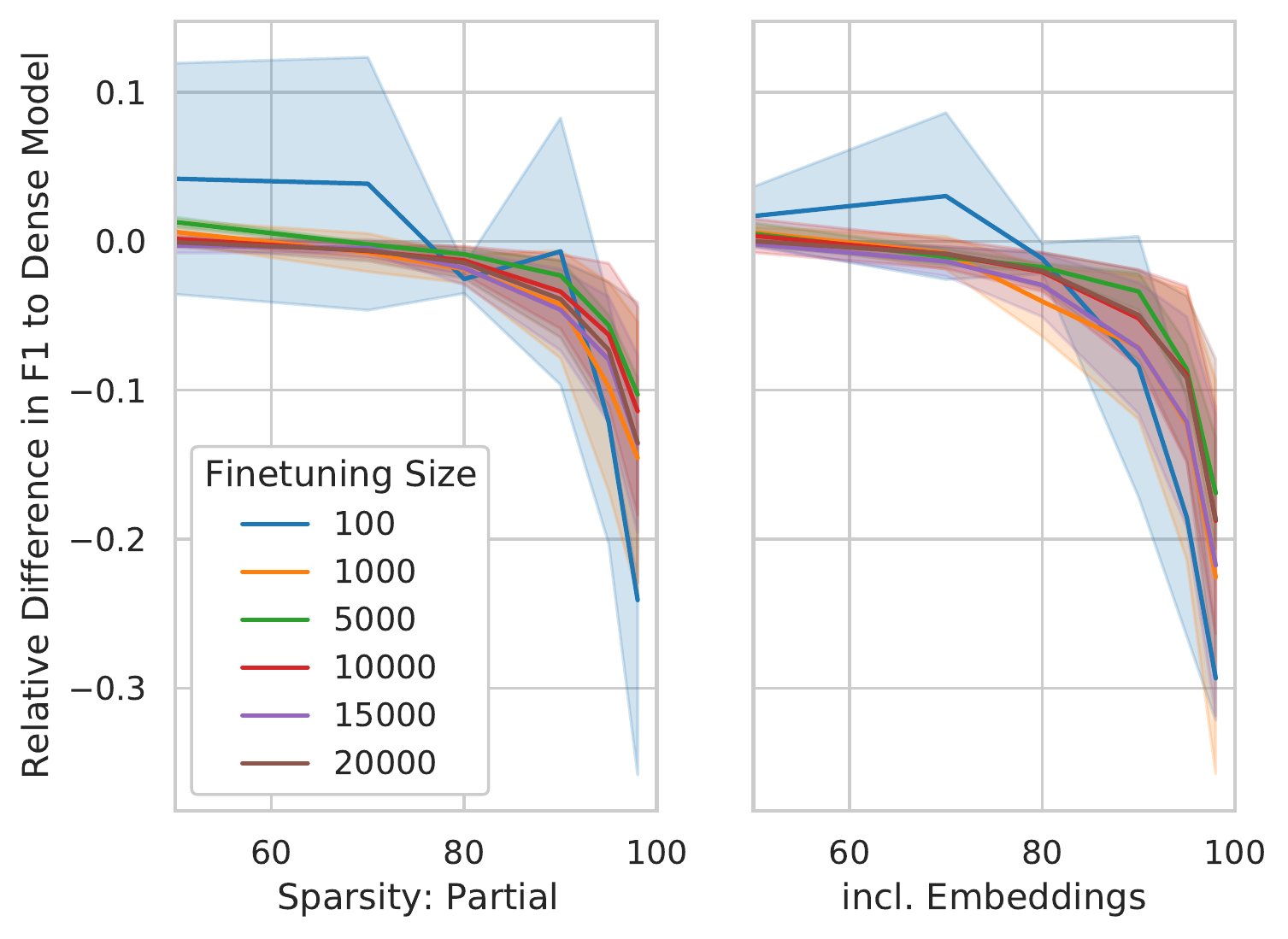}
    \caption{\textbf{Dense vs Sparse:} Mean relative difference in F1 for sparse multilingual models compared to the dense model. Results are averaged for languages grouped according to fine-tuning size. }
    \label{fig:sel_rel_pruning}
\end{figure}

\section{Results and Discussion}
% We investigate the performance of the fine-tuned models from the perspective of multilinguality (\S~\ref{sec:multilingual}), language groups (\S~\ref{sec:languages}) and robustness (\S~\ref{sec:robustness}),
%We start with a set of preliminary ablations to contextualize the experiments before diving into the effects of pruning, 
% highlighting selected findings.
% Detailed results for all language and pruning condition are provided in App.~\ref{app:results} in Tables~\ref{tab:multilingual_without_embeddings} and \ref{tab:multilingual_with_embeddings}.

%\julia{TODO: all: structure according to interesting bits}

%\julia{TODO: all: we need explanations for these results and put them in relation to prior work}

%\julia{TODO: bring in memorization and overfitting}

%Consistent with prior work \citep{}, 
%we find that all languages except for Afrikaans (\texttt{af}) ($-0.003$) benefit from multilingual fine-tuning in comparison to monolingual fine-tuning.
%, with an average difference of $+0.03$ F1, and a maximum of $+0.21$ for Javanese (\texttt{jv}). 

% \paragraph{Cross-lingual transfer also under compression}

\begin{figure*}[ht]
    \centering
    \includegraphics[width=\textwidth]{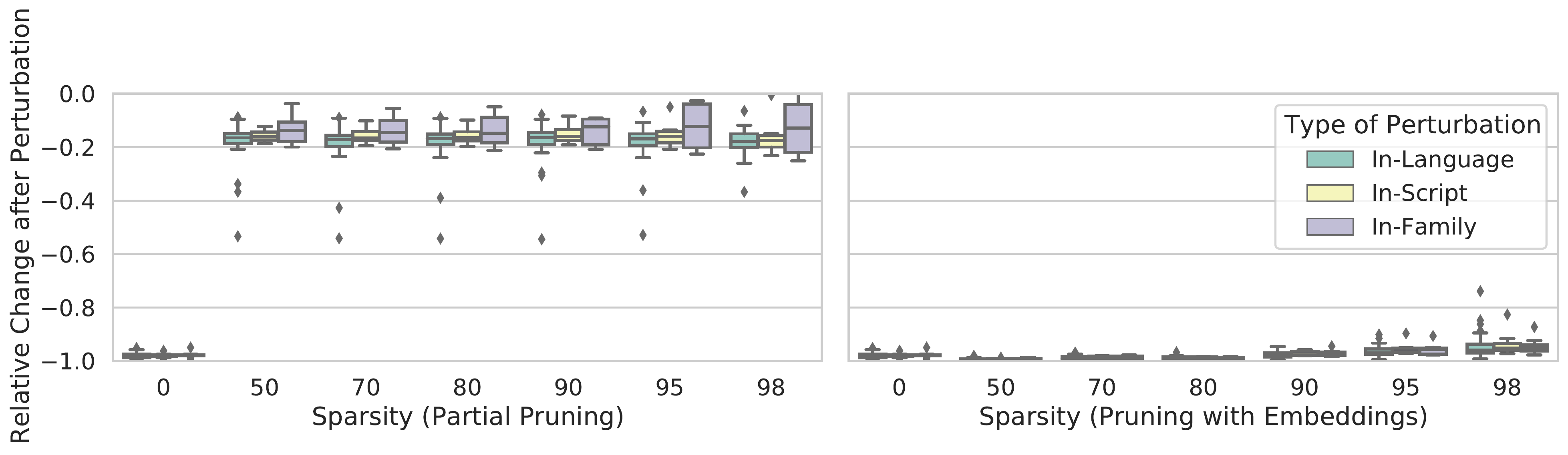}
    \caption{\textbf{Regular vs Perturbed:} We show the aggregated results across all languages after perturbation at different sparsity levels. Without pruning, the model performs poorly, which is overcome by partial pruning, but not pruning with embeddings. The relative performance drop is consistent across all pruning levels above 0.}
    \label{fig:perturbations}
\end{figure*}

\subsection{Multilingual vs. Monolingual}\label{sec:multilingual}
Corroborating prior work on multilingual NER~\citep{hu2020xtreme,adelani-etal-2021-masakhaner}, we find that the multilingual setting generally outperforms the monolingual one. Lower-resource languages tend to benefit more from crosslingual transfer.\footnote{Kappa's $\tau=0.39$ between multilingual gain and fine-tuning size for dense models.}
%The fine-tuning size negatively correlates with the benefits of multilingual training ($\tau=-0.39$), meaning that lower-resource languages tend to benefit more.
%This reconfirms the motivation for multilingual modeling.
% Our experimental settings also allows us to explore whether this is remains the case as sparsity is introduced. 
We find that this finding holds under sparsity -- multilingual models achieve higher F1 than monolingual models not only in the dense setting, but \emph{across all sparsity levels}, as shown in Figure~\ref{fig:multilingual}.\footnote{The three exceptions are Afrikaans (\texttt{af}) at 0\% sparsity ($-0.003$), Hindi (\texttt{hi}) at 50\% sparsity ($-0.01$) for partial pruning, and Yoruba (\texttt{yo}) at 98\% sparsity ($-0.07$) for pruning including the embeddings.} At high sparsity levels, the loss in quality that is generally incurred is considerably lower for multilingual models. This suggests that when high levels of compression are necessary (e.g. for inference efficiency needs), \textit{multilingual training should be preferred to monolingual training}, as it could help offset some of the erosion in the performance caused by the compression. Thus, we conclude that the benefits of \textit{cross-lingual transfer are not inhibited by pruning}, and perhaps are even \textit{more pronounced at a lower capacity}~\citep{dufter-schutze-2020-identifying} for certain languages.
% for, e.g., deployment under hard resource constraints.

%\paragraph{Multilinguality improves robustness}
%The improvements of multilingual models over monolingual models are maintained under the perturbed test scenario, as seen in Figure~\ref{fig:multilingual}. When embeddings are pruned as well, both models largely fail at this task, as observed above.

\subsection{Impact of pruning across languages}\label{sec:languages}
Figure~\ref{fig:sel_rel_pruning} displays the relative differences in F1 score between dense and sparse models across languages, grouped according to fine-tuning size.\footnote{A value of $-0.1$ means that this sparse model reaches $90\%$ quality of the dense model, averaged across the languages within the same size bucket.} 
%We choose this grouping as it showed more distinct characteristics than grouping by language family or script (see Figure~\ref{fig:groups} in Appendix~\ref{app:diagrams}).\julia{revisit this: perhaps show all three groupings? but only unperturbed, and only without embeddings.}
%In the regular testing condition, 
%(top row diagrams in Figure~\ref{fig:abs_pruning}), 
At moderate sparsity levels ($50\%$--$70\%$), \texttt{partial} pruning surprisingly improves over the dense models, in particular those with less fine-tuning data. The majority of languages (26 out of 40) \emph{benefit from moderate pruning} and yield slightly higher F1 with pruning than without. All three datasets with only 100 fine-tuning examples (\texttt{yo}, \texttt{my}, \texttt{jv}) benefit. This suggests that moderate pruning may benefit low-resource datasets when introduced during a finetuning regime. However, at high sparsity levels ($70\%$--$98\%$), the findings reverse. Those languages that have a lower frequency of representation in the finetuning dataset incur the highest absolute and relative loss in quality. We can observe the same trend when grouping languages according to their family or script, respectively (see Fig.~\ref{fig:comp_rel_pruning_family} and~\ref{fig:comp_rel_pruning_script} in App.~\ref{app:diagrams}). The groups that start with the lowest average performance under the dense model, also suffer the most under extreme sparsity.

In conclusion, \textit{moderate pruning levels should be explored for low-resource languages} since they may benefit such languages.
This is especially important since models for low-resourced languages are often deployed in resource-constrained environments§ \cite{ahia-ogueji-towards,nekoto-etal-2020-participatory,ahia-etal-2021-low-resource,ogueji-etal-2021-small,mohammadshahi2022small}. Also, since \textit{high sparsity levels reinforce existing disparities} (as measured by model performance and data availability) between languages and language groups, it is imperative that \textit{practitioners pay attention to possible disparities when sparsifying models}.

\subsection{How does pruning impact robustness?}\label{sec:robustness}
% \paragraph{Overcoming overfitting}
Figure~\ref{fig:perturbations} shows the relative performance on the perturbed sets as a fraction of the corresponding unperturbed performance.
Across all perturbation types, the dense model performs poorly, indicating that the model may have overfit to typical entities and the semantic context that appear in the training corpora.\footnote{Fig~\ref{fig:entity_overlap} shows that entity overlap between train and test set and model performance are correlated. This is particularly obvious for the highest (e.g., \textit{(bn, ur, ms)}) and lowest performing languages (e.g., \textit{(my, yo, jv)}). This may explain the poor performance of dense models on the perturbed test sets.} Surprisingly, \emph{partial pruning at any level} (shown left) improves upon the performance of the dense model. This finding disagrees with some prior works ~\citep{Du2021WhatDC,2019shooker,sehwag2019} which find sparsity erodes different measures of robustness. However, the finding agrees with some other works. For example, \citet{xu-etal-2021-beyond} found that pruning and post-training quantization improve BERT models’ robustness to adversarial examples. Furthermore, \citet{ahia-etal-2021-low-resource} find that magnitude pruning improves model robustness to out-of-distribution shifts in machine translation. Despite the contradictions, our work represents an important step in understanding the impact of pruning on robustness, especially since we are one of the firsts to explore it multilingually. Interestingly, our findings are consistent across all perturbation types as their scope increases from languages (\texttt{in-language}) to scripts (\texttt{in-script}) and families (\texttt{in-family}). This suggests that sparsity can be explored as an avenue to improve robustness as has been explored in previous works \cite{xu-etal-2021-beyond,ahia-etal-2021-low-resource}.

However, pruning the embeddings makes a crucial difference for the perturbed test cases. While pruning the embeddings does not matter for regular test set (see Figure~\ref{fig:sel_rel_pruning}), we observe the same severe drop in performance on the perturbed test-set as for the dense model. This suggests that including \emph{model embeddings when pruning sharply erodes performance} on out-of-distribution rare artefacts, prompting a closer look into what is pruned in the embedding space and the \textit{potential impact of sparsifying different parts of a model}.

\section{Conclusion}
This work investigates the effects of compression on multilingual pre-trained language models during fine-tuning. Our analysis revealed several intriguing properties of pruning that should inform future work in this direction: (1) Pruning dense layers up to $\sim70\%$ may improve quality for low-frequency examples in the data and enhance model robustness. (2) The decision to prune embeddings may have critical impact on model robustness to out-of-distribution performance. (3) While low-performing languages benefit from moderate pruning, they are disproportionately harmed when pruning more aggressively. Based on these intriguing properties, we also make several recommendations to machine learning practitioners.
%Sparsifying pre-trained multilingual models without embeddings could be a valuable step toward improving model robustness during fine-tuning.

\section*{Limitations}
%\sara{kelechi, please add text about limitations here}
%\paragraph{Experimental Scope}
We detail the following potential limitations of our work:

\paragraph{Noisy dataset:} \citet{lignos-etal-2022-toward} shed light on several quality issues of the \wikiann dataset that we are treating as a gold standard. Our results might thus not adequately reflect NER performance that can be achieved with cleaner and human-annotated datasets, such as the MasakhaNER~\citep{adelani-etal-2021-masakhaner} or SADiLaR~\citep{eiselen-2016-government}. Since the perturbations are based on the \wikiann labels, we might be amplifying the existing label noise for the perturbed test sets and as a result underestimate model quality on clean perturbed examples. We try to combat the randomness by averaging results across three separate runs, but any issues intrinsic to \wikiann will likely impact all three.
    
\paragraph{Other Multilingual Models and Downstream tasks: } Multilingual pre-trained models such as XLM-R~\citep{conneau-etal-2020-unsupervised} might yield a better performance or show slightly different trends across languages~\citep{adelani-etal-2021-masakhaner}. Other downstream tasks, especially generation tasks, might tolerate different levels of sparsity, and also show different crosslingual transfer capabilities~\citep{wu-dredze-2019-beto,hu2020xtreme}. However, since fine-grained prior results on the same WikiAnn splits were not available to us, we restricted the analysis to mBERT where we could verify that we can replicate the results reported by XTREME. 
    
\paragraph{Evaluation metrics:} We use F1 as the sole evaluation metric and trust it to reflect quality adequately across languages. Human evaluation and the use of qualitative evaluation metrics might reflect the quality for individual languages better.
    
\paragraph{Unknown factors influencing performance:} The absolute performance for a given language can be influenced by many factors including size, family and script, relatedness to other languages, and the inherent difficulty of the NER task and the evaluation examples, as studied in related works~\citep[e.g.,][]{pires-etal-2019-multilingual, wu-dredze-2020-languages,shaffer-2021-language-clustering,adelani-etal-2021-masakhaner,muller-etal-2021-unseen,deshpande2022when}. As a result, it is impossible to identify the exact cause for all our observations and we have to partially rely on correlational analyses.

\newpage
\bibliography{anthology,custom}

\begin{thebibliography}{71}
\expandafter\ifx\csname natexlab\endcsname\relax\def\natexlab#1{#1}\fi

\bibitem[{ab~Tessera et~al.(2021)ab~Tessera, Hooker, and
  Rosman}]{tessera2021gradients}
Kale ab~Tessera, Sara Hooker, and Benjamin Rosman. 2021.
\newblock \href {http://arxiv.org/abs/2102.01670} {Keep the gradients flowing:
  Using gradient flow to study sparse network optimization}.

\bibitem[{Abdaoui et~al.(2020)Abdaoui, Pradel, and
  Sigel}]{abdaoui-etal-2020-load}
Amine Abdaoui, Camille Pradel, and Gr{\'e}goire Sigel. 2020.
\newblock \href {https://doi.org/10.18653/v1/2020.sustainlp-1.16} {Load what
  you need: Smaller versions of mutililingual {BERT}}.
\newblock In \emph{Proceedings of SustaiNLP: Workshop on Simple and Efficient
  Natural Language Processing}, pages 119--123, Online. Association for
  Computational Linguistics.

\bibitem[{Adelani et~al.(2021)Adelani, Abbott, Neubig, D{'}souza, Kreutzer,
  Lignos, Palen-Michel, Buzaaba, Rijhwani, Ruder, Mayhew, Azime, Muhammad,
  Emezue, Nakatumba-Nabende, Ogayo, Anuoluwapo, Gitau, Mbaye, Alabi, Yimam,
  Gwadabe, Ezeani, Niyongabo, Mukiibi, Otiende, Orife, David, Ngom, Adewumi,
  Rayson, Adeyemi, Muriuki, Anebi, Chukwuneke, Odu, Wairagala, Oyerinde, Siro,
  Bateesa, Oloyede, Wambui, Akinode, Nabagereka, Katusiime, Awokoya, MBOUP,
  Gebreyohannes, Tilaye, Nwaike, Wolde, Faye, Sibanda, Ahia, Dossou, Ogueji,
  DIOP, Diallo, Akinfaderin, Marengereke, and
  Osei}]{adelani-etal-2021-masakhaner}
David~Ifeoluwa Adelani, Jade Abbott, Graham Neubig, Daniel D{'}souza, Julia
  Kreutzer, Constantine Lignos, Chester Palen-Michel, Happy Buzaaba, Shruti
  Rijhwani, Sebastian Ruder, Stephen Mayhew, Israel~Abebe Azime, Shamsuddeen~H.
  Muhammad, Chris~Chinenye Emezue, Joyce Nakatumba-Nabende, Perez Ogayo, Aremu
  Anuoluwapo, Catherine Gitau, Derguene Mbaye, Jesujoba Alabi, Seid~Muhie
  Yimam, Tajuddeen~Rabiu Gwadabe, Ignatius Ezeani, Rubungo~Andre Niyongabo,
  Jonathan Mukiibi, Verrah Otiende, Iroro Orife, Davis David, Samba Ngom, Tosin
  Adewumi, Paul Rayson, Mofetoluwa Adeyemi, Gerald Muriuki, Emmanuel Anebi,
  Chiamaka Chukwuneke, Nkiruka Odu, Eric~Peter Wairagala, Samuel Oyerinde,
  Clemencia Siro, Tobius~Saul Bateesa, Temilola Oloyede, Yvonne Wambui, Victor
  Akinode, Deborah Nabagereka, Maurice Katusiime, Ayodele Awokoya, Mouhamadane
  MBOUP, Dibora Gebreyohannes, Henok Tilaye, Kelechi Nwaike, Degaga Wolde,
  Abdoulaye Faye, Blessing Sibanda, Orevaoghene Ahia, Bonaventure F.~P. Dossou,
  Kelechi Ogueji, Thierno~Ibrahima DIOP, Abdoulaye Diallo, Adewale Akinfaderin,
  Tendai Marengereke, and Salomey Osei. 2021.
\newblock \href {https://doi.org/10.1162/tacl_a_00416} {{M}asakha{NER}: Named
  entity recognition for {A}frican languages}.
\newblock \emph{Transactions of the Association for Computational Linguistics},
  9:1116--1131.

\bibitem[{Ahia et~al.(2021)Ahia, Kreutzer, and
  Hooker}]{ahia-etal-2021-low-resource}
Orevaoghene Ahia, Julia Kreutzer, and Sara Hooker. 2021.
\newblock \href {https://doi.org/10.18653/v1/2021.findings-emnlp.282} {The
  low-resource double bind: An empirical study of pruning for low-resource
  machine translation}.
\newblock In \emph{Findings of the Association for Computational Linguistics:
  EMNLP 2021}, pages 3316--3333, Punta Cana, Dominican Republic. Association
  for Computational Linguistics.

\bibitem[{Ahia and Ogueji(2020)}]{ahia-ogueji-towards}
Orevaoghene Ahia and Kelechi Ogueji. 2020.
\newblock Towards supervised and unsupervised neural machine translation
  baselines for nigerian pidgin.
\newblock \emph{ArXiv}, abs/2003.12660.

\bibitem[{Ansell et~al.(2022)Ansell, Ponti, Korhonen, and
  Vuli{\'c}}]{ansell-etal-2022-composable}
Alan Ansell, Edoardo Ponti, Anna Korhonen, and Ivan Vuli{\'c}. 2022.
\newblock \href {https://doi.org/10.18653/v1/2022.acl-long.125} {Composable
  sparse fine-tuning for cross-lingual transfer}.
\newblock In \emph{Proceedings of the 60th Annual Meeting of the Association
  for Computational Linguistics (Volume 1: Long Papers)}, pages 1778--1796,
  Dublin, Ireland. Association for Computational Linguistics.

\bibitem[{Artetxe and Schwenk(2019)}]{Artetxe2019MassivelyMS}
Mikel Artetxe and Holger Schwenk. 2019.
\newblock Massively multilingual sentence embeddings for zero-shot
  cross-lingual transfer and beyond.
\newblock \emph{Transactions of the Association for Computational Linguistics},
  7:597--610.

\bibitem[{Bai et~al.(2020)Bai, Zhang, Hou, Shang, Jin, Jiang, Liu, Lyu, and
  King}]{Bai2020BinaryBERTPT}
Haoli Bai, Wei Zhang, Lu~Hou, Lifeng Shang, Jing Jin, X.~Jiang, Qun Liu,
  Michael~R. Lyu, and Irwin King. 2020.
\newblock Binarybert: Pushing the limit of bert quantization.
\newblock \emph{ArXiv}, abs/2012.15701.

\bibitem[{Brown et~al.(2020)Brown, Mann, Ryder, Subbiah, Kaplan, Dhariwal,
  Neelakantan, Shyam, Sastry, Askell, Agarwal, Herbert-Voss, Krueger, Henighan,
  Child, Ramesh, Ziegler, Wu, Winter, Hesse, Chen, Sigler, Litwin, Gray, Chess,
  Clark, Berner, McCandlish, Radford, Sutskever, and
  Amodei}]{Brown2020LanguageMA}
Tom~B. Brown, Benjamin Mann, Nick Ryder, Melanie Subbiah, Jared Kaplan,
  Prafulla Dhariwal, Arvind Neelakantan, Pranav Shyam, Girish Sastry, Amanda
  Askell, Sandhini Agarwal, Ariel Herbert-Voss, Gretchen Krueger, T.~J.
  Henighan, Rewon Child, Aditya Ramesh, Daniel~M. Ziegler, Jeff Wu, Clemens
  Winter, Christopher Hesse, Mark Chen, Eric Sigler, Mateusz Litwin, Scott
  Gray, Benjamin Chess, Jack Clark, Christopher Berner, Sam McCandlish, Alec
  Radford, Ilya Sutskever, and Dario Amodei. 2020.
\newblock Language models are few-shot learners.
\newblock \emph{ArXiv}, abs/2005.14165.

\bibitem[{Budhraja et~al.(2021)Budhraja, Pande, Kumar, and
  Khapra}]{Budhraja2021OnTP}
Aakriti Budhraja, Madhura Pande, Pratyush Kumar, and Mitesh~M. Khapra. 2021.
\newblock On the prunability of attention heads in multilingual bert.
\newblock \emph{ArXiv}, abs/2109.12683.

\bibitem[{Budhraja et~al.(2020)Budhraja, Pande, Nema, Kumar, and
  Khapra}]{budhraja-etal-2020-weak}
Aakriti Budhraja, Madhura Pande, Preksha Nema, Pratyush Kumar, and Mitesh~M.
  Khapra. 2020.
\newblock \href {https://doi.org/10.18653/v1/2020.emnlp-main.260} {On the weak
  link between importance and prunability of attention heads}.
\newblock In \emph{Proceedings of the 2020 Conference on Empirical Methods in
  Natural Language Processing (EMNLP)}, pages 3230--3235, Online. Association
  for Computational Linguistics.

\bibitem[{Chen et~al.(2021)Chen, Cheng, Wang, Gan, Wang, and jing
  Liu}]{Chen2021EarlyBERTEB}
Xiao-Han Chen, Yu~Cheng, Shuohang Wang, Zhe Gan, Zhangyang Wang, and Jing jing
  Liu. 2021.
\newblock Earlybert: Efficient bert training via early-bird lottery tickets.
\newblock \emph{ArXiv}, abs/2101.00063.

\bibitem[{Chi et~al.(2020)Chi, Hewitt, and Manning}]{chi-etal-2020-finding}
Ethan~A. Chi, John Hewitt, and Christopher~D. Manning. 2020.
\newblock \href {https://doi.org/10.18653/v1/2020.acl-main.493} {Finding
  universal grammatical relations in multilingual {BERT}}.
\newblock In \emph{Proceedings of the 58th Annual Meeting of the Association
  for Computational Linguistics}, pages 5564--5577, Online. Association for
  Computational Linguistics.

\bibitem[{Chowdhery et~al.(2022)Chowdhery, Narang, Devlin, Bosma, Mishra,
  Roberts, Barham, Chung, Sutton, Gehrmann, Schuh, Shi, Tsvyashchenko, Maynez,
  Rao, Barnes, Tay, Shazeer, Prabhakaran, Reif, Du, Hutchinson, Pope, Bradbury,
  Austin, Isard, Gur-Ari, Yin, Duke, Levskaya, Ghemawat, Dev, Michalewski,
  Garc{\'i}a, Misra, Robinson, Fedus, Zhou, Ippolito, Luan, Lim, Zoph,
  Spiridonov, Sepassi, Dohan, Agrawal, Omernick, Dai, Pillai, Pellat,
  Lewkowycz, Moreira, Child, Polozov, Lee, Zhou, Wang, Saeta, Diaz, Firat,
  Catasta, Wei, Meier-Hellstern, Eck, Dean, Petrov, and
  Fiedel}]{Chowdhery2022PaLMSL}
Aakanksha Chowdhery, Sharan Narang, Jacob Devlin, Maarten Bosma, Gaurav Mishra,
  Adam Roberts, Paul Barham, Hyung~Won Chung, Charles Sutton, Sebastian
  Gehrmann, Parker Schuh, Kensen Shi, Sasha Tsvyashchenko, Joshua Maynez,
  Abhishek~B Rao, Parker Barnes, Yi~Tay, Noam~M. Shazeer, Vinodkumar
  Prabhakaran, Emily Reif, Nan Du, Benton~C. Hutchinson, Reiner Pope, James
  Bradbury, Jacob Austin, Michael Isard, Guy Gur-Ari, Pengcheng Yin, Toju Duke,
  Anselm Levskaya, Sanjay Ghemawat, Sunipa Dev, Henryk Michalewski, Xavier
  Garc{\'i}a, Vedant Misra, Kevin Robinson, Liam Fedus, Denny Zhou, Daphne
  Ippolito, David Luan, Hyeontaek Lim, Barret Zoph, Alexander Spiridonov, Ryan
  Sepassi, David Dohan, Shivani Agrawal, Mark Omernick, Andrew~M. Dai,
  Thanumalayan~Sankaranarayana Pillai, Marie Pellat, Aitor Lewkowycz,
  Erica~Oliveira Moreira, Rewon Child, Oleksandr Polozov, Katherine Lee,
  Zongwei Zhou, Xuezhi Wang, Brennan Saeta, Mark Diaz, Orhan Firat, Michele
  Catasta, Jason Wei, Kathleen~S. Meier-Hellstern, Douglas Eck, Jeff Dean, Slav
  Petrov, and Noah Fiedel. 2022.
\newblock Palm: Scaling language modeling with pathways.
\newblock \emph{ArXiv}, abs/2204.02311.

\bibitem[{Conneau et~al.(2020)Conneau, Khandelwal, Goyal, Chaudhary, Wenzek,
  Guzm{\'a}n, Grave, Ott, Zettlemoyer, and
  Stoyanov}]{conneau-etal-2020-unsupervised}
Alexis Conneau, Kartikay Khandelwal, Naman Goyal, Vishrav Chaudhary, Guillaume
  Wenzek, Francisco Guzm{\'a}n, Edouard Grave, Myle Ott, Luke Zettlemoyer, and
  Veselin Stoyanov. 2020.
\newblock \href {https://doi.org/10.18653/v1/2020.acl-main.747} {Unsupervised
  cross-lingual representation learning at scale}.
\newblock In \emph{Proceedings of the 58th Annual Meeting of the Association
  for Computational Linguistics}, pages 8440--8451, Online. Association for
  Computational Linguistics.

\bibitem[{Conneau and Lample(2019)}]{NEURIPS2019_c04c19c2}
Alexis Conneau and Guillaume Lample. 2019.
\newblock \href
  {https://proceedings.neurips.cc/paper/2019/file/c04c19c2c2474dbf5f7ac4372c5b9af1-Paper.pdf}
  {Cross-lingual language model pretraining}.
\newblock In \emph{Advances in Neural Information Processing Systems},
  volume~32. Curran Associates, Inc.

\bibitem[{Dai and Adel(2020)}]{dai-adel-2020-analysis}
Xiang Dai and Heike Adel. 2020.
\newblock \href {https://doi.org/10.18653/v1/2020.coling-main.343} {An analysis
  of simple data augmentation for named entity recognition}.
\newblock In \emph{Proceedings of the 28th International Conference on
  Computational Linguistics}, pages 3861--3867, Barcelona, Spain (Online).
  International Committee on Computational Linguistics.

\bibitem[{Deshpande et~al.(2021)Deshpande, Talukdar, and
  Narasimhan}]{deshpande2022when}
Ameet Deshpande, Partha Talukdar, and Karthik Narasimhan. 2021.
\newblock \href {http://arxiv.org/abs/2110.14782} {When is {BERT} multilingual?
  isolating crucial ingredients for cross-lingual transfer}.
\newblock \emph{CoRR}, abs/2110.14782.

\bibitem[{Devlin et~al.(2019)Devlin, Chang, Lee, and
  Toutanova}]{devlin-etal-2019-bert}
Jacob Devlin, Ming-Wei Chang, Kenton Lee, and Kristina Toutanova. 2019.
\newblock \href {https://doi.org/10.18653/v1/N19-1423} {{BERT}: Pre-training of
  deep bidirectional transformers for language understanding}.
\newblock In \emph{Proceedings of the 2019 Conference of the North {A}merican
  Chapter of the Association for Computational Linguistics: Human Language
  Technologies, Volume 1 (Long and Short Papers)}, pages 4171--4186,
  Minneapolis, Minnesota. Association for Computational Linguistics.

\bibitem[{Dhole et~al.(2021)Dhole, Gangal, Gehrmann, Gupta, Li, Mahamood,
  Mahendiran, Mille, Srivastava, Tan et~al.}]{dhole2021nl}
Kaustubh~D Dhole, Varun Gangal, Sebastian Gehrmann, Aadesh Gupta, Zhenhao Li,
  Saad Mahamood, Abinaya Mahendiran, Simon Mille, Ashish Srivastava, Samson
  Tan, et~al. 2021.
\newblock Nl-augmenter: A framework for task-sensitive natural language
  augmentation.
\newblock \emph{arXiv preprint arXiv:2112.02721}.

\bibitem[{Du et~al.(2021{\natexlab{a}})Du, Mukherjee, Cheng, Shokouhi, Hu, and
  Awadallah}]{Du2021WhatDC}
Mengnan Du, Subhabrata Mukherjee, Yu~Cheng, Milad Shokouhi, Xia Hu, and
  Ahmed~Hassan Awadallah. 2021{\natexlab{a}}.
\newblock \href {http://arxiv.org/abs/2110.08419} {What do compressed large
  language models forget? robustness challenges in model compression}.
\newblock \emph{CoRR}, abs/2110.08419.

\bibitem[{Du et~al.(2021{\natexlab{b}})Du, Mukherjee, Cheng, Shokouhi, Hu, and
  Awadallah}]{mengnan-forget}
Mengnan Du, Subhabrata Mukherjee, Yu~Cheng, Milad Shokouhi, Xia Hu, and
  Ahmed~Hassan Awadallah. 2021{\natexlab{b}}.
\newblock What do compressed large language models forget? robustness
  challenges in model compression.
\newblock \emph{ArXiv}, abs/2110.08419.

\bibitem[{Dufter and Sch{\"u}tze(2020)}]{dufter-schutze-2020-identifying}
Philipp Dufter and Hinrich Sch{\"u}tze. 2020.
\newblock \href {https://doi.org/10.18653/v1/2020.emnlp-main.358} {Identifying
  elements essential for {BERT}{'}s multilinguality}.
\newblock In \emph{Proceedings of the 2020 Conference on Empirical Methods in
  Natural Language Processing (EMNLP)}, pages 4423--4437, Online. Association
  for Computational Linguistics.

\bibitem[{Eiselen(2016)}]{eiselen-2016-government}
Roald Eiselen. 2016.
\newblock \href {https://aclanthology.org/L16-1533} {Government domain named
  entity recognition for {S}outh {A}frican languages}.
\newblock In \emph{Proceedings of the Tenth International Conference on
  Language Resources and Evaluation ({LREC}'16)}, pages 3344--3348,
  Portoro{\v{z}}, Slovenia. European Language Resources Association (ELRA).

\bibitem[{Gale et~al.(2019)Gale, Elsen, and Hooker}]{gale2019}
Trevor Gale, Erich Elsen, and Sara Hooker. 2019.
\newblock \href {http://arxiv.org/abs/1902.09574} {The state of sparsity in
  deep neural networks}.
\newblock \emph{CoRR}, abs/1902.09574.

\bibitem[{{Gale} et~al.(2019){Gale}, {Elsen}, and
  {Hooker}}]{2019arXiv190209574G}
Trevor {Gale}, Erich {Elsen}, and Sara {Hooker}. 2019.
\newblock \href {http://arxiv.org/abs/1902.09574} {{The State of Sparsity in
  Deep Neural Networks}}.
\newblock \emph{arXiv e-prints}, page arXiv:1902.09574.

\bibitem[{Ganesh et~al.(2021)Ganesh, Chen, Lou, Khan, Yang, Sajjad, Nakov,
  Chen, and Winslett}]{ganesh-etal-2021-compressing}
Prakhar Ganesh, Yao Chen, Xin Lou, Mohammad~Ali Khan, Yin Yang, Hassan Sajjad,
  Preslav Nakov, Deming Chen, and Marianne Winslett. 2021.
\newblock \href {https://doi.org/10.1162/tacl_a_00413} {Compressing large-scale
  transformer-based models: A case study on {BERT}}.
\newblock \emph{Transactions of the Association for Computational Linguistics},
  9:1061--1080.

\bibitem[{Gordon et~al.(2020)Gordon, Duh, and
  Andrews}]{gordon-etal-2020-compressing}
Mitchell Gordon, Kevin Duh, and Nicholas Andrews. 2020.
\newblock \href {https://doi.org/10.18653/v1/2020.repl4nlp-1.18} {Compressing
  {BERT}: Studying the effects of weight pruning on transfer learning}.
\newblock In \emph{Proceedings of the 5th Workshop on Representation Learning
  for NLP}, pages 143--155, Online. Association for Computational Linguistics.

\bibitem[{Goyal et~al.(2020)Goyal, Choudhury, Raje, Chakaravarthy, Sabharwal,
  and Verma}]{Goyal2020PoWERBERTAB}
Saurabh Goyal, Anamitra~R. Choudhury, Saurabh Raje, Venkatesan~T.
  Chakaravarthy, Yogish Sabharwal, and Ashish Verma. 2020.
\newblock Power-bert: Accelerating bert inference via progressive word-vector
  elimination.
\newblock In \emph{ICML}.

\bibitem[{Han et~al.(2016)Han, Mao, and Dally}]{han2016deep}
Song Han, Huizi Mao, and William~J. Dally. 2016.
\newblock \href {http://arxiv.org/abs/1510.00149} {Deep compression:
  Compressing deep neural networks with pruning, trained quantization and
  huffman coding}.

\bibitem[{Han et~al.(2015)Han, Pool, Tran, and Dally}]{Han2015}
Song Han, Jeff Pool, John Tran, and William~J. Dally. 2015.
\newblock \href {http://dl.acm.org/citation.cfm?id=2969239.2969366} {Learning
  both weights and connections for efficient neural networks}.
\newblock In \emph{Proceedings of the 28th International Conference on Neural
  Information Processing Systems - Volume 1}, NeurIPS'15, pages 1135--1143,
  Cambridge, MA, USA. MIT Press.

\bibitem[{{Hooker} et~al.(2019){Hooker}, {Courville}, {Clark}, {Dauphin}, and
  {Frome}}]{2019shooker}
Sara {Hooker}, Aaron {Courville}, Gregory {Clark}, Yann {Dauphin}, and Andrea
  {Frome}. 2019.
\newblock \href {http://arxiv.org/abs/1911.05248} {{What Do Compressed Deep
  Neural Networks Forget?}}
\newblock \emph{arXiv e-prints}, page arXiv:1911.05248.

\bibitem[{Hooker et~al.(2020)Hooker, Moorosi, Clark, Bengio, and
  Denton}]{2020hooker}
Sara Hooker, Nyalleng Moorosi, Gregory Clark, Samy Bengio, and Emily Denton.
  2020.
\newblock \href {https://doi.org/10.48550/ARXIV.2010.03058} {Characterising
  bias in compressed models}.

\bibitem[{Hou et~al.(2020)Hou, Shang, Jiang, and Liu}]{Hou2020DynaBERTDB}
Lu~Hou, Lifeng Shang, X.~Jiang, and Qun Liu. 2020.
\newblock Dynabert: Dynamic bert with adaptive width and depth.
\newblock \emph{ArXiv}, abs/2004.04037.

\bibitem[{Hu et~al.(2020)Hu, Ruder, Siddhant, Neubig, Firat, and
  Johnson}]{hu2020xtreme}
Junjie Hu, Sebastian Ruder, Aditya Siddhant, Graham Neubig, Orhan Firat, and
  Melvin Johnson. 2020.
\newblock \href {http://arxiv.org/abs/2003.11080} {Xtreme: A massively
  multilingual multi-task benchmark for evaluating cross-lingual
  generalization}.
\newblock \emph{CoRR}, abs/2003.11080.

\bibitem[{Lagunas et~al.(2021)Lagunas, Charlaix, Sanh, and
  Rush}]{lagunas-etal-2021-block}
Fran{\c{c}}ois Lagunas, Ella Charlaix, Victor Sanh, and Alexander Rush. 2021.
\newblock \href {https://doi.org/10.18653/v1/2021.emnlp-main.829} {Block
  pruning for faster transformers}.
\newblock In \emph{Proceedings of the 2021 Conference on Empirical Methods in
  Natural Language Processing}, pages 10619--10629, Online and Punta Cana,
  Dominican Republic. Association for Computational Linguistics.

\bibitem[{Li et~al.(2020)Li, Wang, Liu, Du, Xiao, Zhang, and
  Zhu}]{Li2020LearningLT}
Bei Li, Ziyang Wang, H.~Liu, Quan Du, Tong Xiao, Chunliang Zhang, and Jingbo
  Zhu. 2020.
\newblock Learning light-weight translation models from deep transformer.
\newblock \emph{ArXiv}, abs/2012.13866.

\bibitem[{Lignos et~al.(2022)Lignos, Holley, Palen-Michel, and
  S{\"a}lev{\"a}}]{lignos-etal-2022-toward}
Constantine Lignos, Nolan Holley, Chester Palen-Michel, and Jonne
  S{\"a}lev{\"a}. 2022.
\newblock \href {https://doi.org/10.18653/v1/2022.findings-acl.44} {Toward more
  meaningful resources for lower-resourced languages}.
\newblock In \emph{Findings of the Association for Computational Linguistics:
  ACL 2022}, pages 523--532, Dublin, Ireland. Association for Computational
  Linguistics.

\bibitem[{Loshchilov and Hutter(2019)}]{Loshchilov2019DecoupledWD}
Ilya Loshchilov and Frank Hutter. 2019.
\newblock Decoupled weight decay regularization.
\newblock In \emph{ICLR}.

\bibitem[{Ma et~al.(2021)Ma, Zhang, Lou, Wang, and
  Vosoughi}]{ma-etal-2021-contributions}
Weicheng Ma, Kai Zhang, Renze Lou, Lili Wang, and Soroush Vosoughi. 2021.
\newblock \href {https://doi.org/10.18653/v1/2021.acl-long.152} {Contributions
  of transformer attention heads in multi- and cross-lingual tasks}.
\newblock In \emph{Proceedings of the 59th Annual Meeting of the Association
  for Computational Linguistics and the 11th International Joint Conference on
  Natural Language Processing (Volume 1: Long Papers)}, pages 1956--1966,
  Online. Association for Computational Linguistics.

\bibitem[{Michel et~al.(2019)Michel, Levy, and Neubig}]{michel2019sixteen}
Paul Michel, Omer Levy, and Graham Neubig. 2019.
\newblock Are sixteen heads really better than one?
\newblock \emph{Advances in neural information processing systems (NeurIPS)},
  32.

\bibitem[{Mohammadshahi et~al.(2022{\natexlab{a}})Mohammadshahi, Nikoulina,
  Berard, Brun, Henderson, and Besacier}]{mohammadshahi2022small}
Alireza Mohammadshahi, Vassilina Nikoulina, Alexandre Berard, Caroline Brun,
  James Henderson, and Laurent Besacier. 2022{\natexlab{a}}.
\newblock Small-100: Introducing shallow multilingual machine translation model
  for low-resource languages.
\newblock \emph{arXiv preprint arXiv:2210.11621}.

\bibitem[{Mohammadshahi et~al.(2022{\natexlab{b}})Mohammadshahi, Nikoulina,
  Berard, Brun, Henderson, and Besacier}]{mohammadshahi2022compressed}
Alireza Mohammadshahi, Vassilina Nikoulina, Alexandre Berard, Caroline Brun,
  James Henderson, and Laurent Besacier. 2022{\natexlab{b}}.
\newblock What do compressed multilingual machine translation models forget?
\newblock \emph{arXiv preprint arXiv:2205.10828}.

\bibitem[{Mukherjee and
  Hassan~Awadallah(2020)}]{mukherjee-hassan-awadallah-2020-xtremedistil}
Subhabrata Mukherjee and Ahmed Hassan~Awadallah. 2020.
\newblock \href {https://doi.org/10.18653/v1/2020.acl-main.202}
  {{X}treme{D}istil: Multi-stage distillation for massive multilingual models}.
\newblock In \emph{Proceedings of the 58th Annual Meeting of the Association
  for Computational Linguistics}, pages 2221--2234, Online. Association for
  Computational Linguistics.

\bibitem[{Muller et~al.(2021)Muller, Anastasopoulos, Sagot, and
  Seddah}]{muller-etal-2021-unseen}
Benjamin Muller, Antonios Anastasopoulos, Beno{\^\i}t Sagot, and Djam{\'e}
  Seddah. 2021.
\newblock \href {https://doi.org/10.18653/v1/2021.naacl-main.38} {When being
  unseen from m{BERT} is just the beginning: Handling new languages with
  multilingual language models}.
\newblock In \emph{Proceedings of the 2021 Conference of the North American
  Chapter of the Association for Computational Linguistics: Human Language
  Technologies}, pages 448--462, Online. Association for Computational
  Linguistics.

\bibitem[{Nakayama(2018)}]{seqeval}
Hiroki Nakayama. 2018.
\newblock \href {https://github.com/chakki-works/seqeval} {{seqeval}: A python
  framework for sequence labeling evaluation}.
\newblock Software available from https://github.com/chakki-works/seqeval.

\bibitem[{Nekoto et~al.(2020)Nekoto, Marivate, Matsila, Fasubaa, Fagbohungbe,
  Akinola, Muhammad, Kabongo~Kabenamualu, Osei, Sackey, Niyongabo, Macharm,
  Ogayo, Ahia, Berhe, Adeyemi, Mokgesi-Selinga, Okegbemi, Martinus, Tajudeen,
  Degila, Ogueji, Siminyu, Kreutzer, Webster, Ali, Abbott, Orife, Ezeani,
  Dangana, Kamper, Elsahar, Duru, Kioko, Espoir, van Biljon, Whitenack,
  Onyefuluchi, Emezue, Dossou, Sibanda, Bassey, Olabiyi, Ramkilowan, {\"O}ktem,
  Akinfaderin, and Bashir}]{nekoto-etal-2020-participatory}
Wilhelmina Nekoto, Vukosi Marivate, Tshinondiwa Matsila, Timi Fasubaa, Taiwo
  Fagbohungbe, Solomon~Oluwole Akinola, Shamsuddeen Muhammad, Salomon
  Kabongo~Kabenamualu, Salomey Osei, Freshia Sackey, Rubungo~Andre Niyongabo,
  Ricky Macharm, Perez Ogayo, Orevaoghene Ahia, Musie~Meressa Berhe, Mofetoluwa
  Adeyemi, Masabata Mokgesi-Selinga, Lawrence Okegbemi, Laura Martinus,
  Kolawole Tajudeen, Kevin Degila, Kelechi Ogueji, Kathleen Siminyu, Julia
  Kreutzer, Jason Webster, Jamiil~Toure Ali, Jade Abbott, Iroro Orife, Ignatius
  Ezeani, Idris~Abdulkadir Dangana, Herman Kamper, Hady Elsahar, Goodness Duru,
  Ghollah Kioko, Murhabazi Espoir, Elan van Biljon, Daniel Whitenack,
  Christopher Onyefuluchi, Chris~Chinenye Emezue, Bonaventure F.~P. Dossou,
  Blessing Sibanda, Blessing Bassey, Ayodele Olabiyi, Arshath Ramkilowan, Alp
  {\"O}ktem, Adewale Akinfaderin, and Abdallah Bashir. 2020.
\newblock \href {https://doi.org/10.18653/v1/2020.findings-emnlp.195}
  {Participatory research for low-resourced machine translation: A case study
  in {A}frican languages}.
\newblock In \emph{Findings of the Association for Computational Linguistics:
  EMNLP 2020}, pages 2144--2160, Online. Association for Computational
  Linguistics.

\bibitem[{Ogueji et~al.(2021)Ogueji, Zhu, and Lin}]{ogueji-etal-2021-small}
Kelechi Ogueji, Yuxin Zhu, and Jimmy Lin. 2021.
\newblock \href {https://doi.org/10.18653/v1/2021.mrl-1.11} {Small data? no
  problem! exploring the viability of pretrained multilingual language models
  for low-resourced languages}.
\newblock In \emph{Proceedings of the 1st Workshop on Multilingual
  Representation Learning}, pages 116--126, Punta Cana, Dominican Republic.
  Association for Computational Linguistics.

\bibitem[{Pan et~al.(2017)Pan, Zhang, May, Nothman, Knight, and
  Ji}]{pan-etal-2017-cross}
Xiaoman Pan, Boliang Zhang, Jonathan May, Joel Nothman, Kevin Knight, and Heng
  Ji. 2017.
\newblock \href {https://doi.org/10.18653/v1/P17-1178} {Cross-lingual name
  tagging and linking for 282 languages}.
\newblock In \emph{Proceedings of the 55th Annual Meeting of the Association
  for Computational Linguistics (Volume 1: Long Papers)}, pages 1946--1958,
  Vancouver, Canada. Association for Computational Linguistics.

\bibitem[{Pires et~al.(2019)Pires, Schlinger, and
  Garrette}]{pires-etal-2019-multilingual}
Telmo Pires, Eva Schlinger, and Dan Garrette. 2019.
\newblock \href {https://doi.org/10.18653/v1/P19-1493} {How multilingual is
  multilingual {BERT}?}
\newblock In \emph{Proceedings of the 57th Annual Meeting of the Association
  for Computational Linguistics}, pages 4996--5001, Florence, Italy.
  Association for Computational Linguistics.

\bibitem[{Pu et~al.(2021)Pu, Chung, Parikh, Gehrmann, and
  Sellam}]{pu-etal-2021-learning}
Amy Pu, Hyung~Won Chung, Ankur Parikh, Sebastian Gehrmann, and Thibault Sellam.
  2021.
\newblock \href {https://doi.org/10.18653/v1/2021.emnlp-main.58} {Learning
  compact metrics for {MT}}.
\newblock In \emph{Proceedings of the 2021 Conference on Empirical Methods in
  Natural Language Processing}, pages 751--762, Online and Punta Cana,
  Dominican Republic. Association for Computational Linguistics.

\bibitem[{Radford et~al.(2019)Radford, Wu, Child, Luan, Amodei, and
  Sutskever}]{Radford2019LanguageMA}
Alec Radford, Jeff Wu, Rewon Child, David Luan, Dario Amodei, and Ilya
  Sutskever. 2019.
\newblock Language models are unsupervised multitask learners.

\bibitem[{Rahimi et~al.(2019)Rahimi, Li, and Cohn}]{rahimi-etal-2019-massively}
Afshin Rahimi, Yuan Li, and Trevor Cohn. 2019.
\newblock \href {https://doi.org/10.18653/v1/P19-1015} {Massively multilingual
  transfer for {NER}}.
\newblock In \emph{Proceedings of the 57th Annual Meeting of the Association
  for Computational Linguistics}, pages 151--164, Florence, Italy. Association
  for Computational Linguistics.

\bibitem[{Ramshaw and Marcus(1995)}]{ramshaw-marcus-1995-text}
Lance Ramshaw and Mitch Marcus. 1995.
\newblock \href {https://aclanthology.org/W95-0107} {Text chunking using
  transformation-based learning}.
\newblock In \emph{Third Workshop on Very Large Corpora}.

\bibitem[{R{\"o}nnqvist et~al.(2019)R{\"o}nnqvist, Kanerva, Salakoski, and
  Ginter}]{ronnqvist-etal-2019-multilingual}
Samuel R{\"o}nnqvist, Jenna Kanerva, Tapio Salakoski, and Filip Ginter. 2019.
\newblock \href {https://aclanthology.org/W19-6204} {Is multilingual {BERT}
  fluent in language generation?}
\newblock In \emph{Proceedings of the First NLPL Workshop on Deep Learning for
  Natural Language Processing}, pages 29--36, Turku, Finland. Link{\"o}ping
  University Electronic Press.

\bibitem[{Sajjad et~al.(2020)Sajjad, Dalvi, Durrani, and
  Nakov}]{Sajjad2020PoorMB}
Hassan Sajjad, Fahim Dalvi, Nadir Durrani, and Preslav Nakov. 2020.
\newblock Poor man's bert: Smaller and faster transformer models.
\newblock \emph{ArXiv}, abs/2004.03844.

\bibitem[{Samala et~al.(2018)Samala, Chan, Hadjiiski, Helvie, Richter, and
  Cha}]{Samala_2018}
Ravi~K Samala, Heang-Ping Chan, Lubomir~M Hadjiiski, Mark~A Helvie, Caleb
  Richter, and Kenny Cha. 2018.
\newblock \href {https://doi.org/10.1088/1361-6560/aabb5b} {Evolutionary
  pruning of transfer learned deep convolutional neural network for breast
  cancer diagnosis in digital breast tomosynthesis}.
\newblock \emph{Physics in Medicine {\&} Biology}, 63(9):095005.

\bibitem[{Sanh et~al.(2019)Sanh, Debut, Chaumond, and
  Wolf}]{sanh2019distilbert}
Victor Sanh, Lysandre Debut, Julien Chaumond, and Thomas Wolf. 2019.
\newblock Distilbert, a distilled version of bert: smaller, faster, cheaper and
  lighter.
\newblock \emph{arXiv preprint arXiv:1910.01108}.

\bibitem[{Sanh et~al.(2020)Sanh, Wolf, and Rush}]{NEURIPS2020_eae15aab}
Victor Sanh, Thomas Wolf, and Alexander Rush. 2020.
\newblock \href
  {https://proceedings.neurips.cc/paper/2020/file/eae15aabaa768ae4a5993a8a4f4fa6e4-Paper.pdf}
  {Movement pruning: Adaptive sparsity by fine-tuning}.
\newblock In \emph{Advances in Neural Information Processing Systems},
  volume~33, pages 20378--20389. Curran Associates, Inc.

\bibitem[{Sehwag et~al.(2019)Sehwag, Wang, Mittal, and Jana}]{sehwag2019}
Vikash Sehwag, Shiqi Wang, Prateek Mittal, and Suman Jana. 2019.
\newblock \href {http://arxiv.org/abs/1906.06110} {Towards compact and robust
  deep neural networks}.
\newblock \emph{CoRR}, abs/1906.06110.

\bibitem[{Shaffer(2021)}]{shaffer-2021-language-clustering}
Kyle Shaffer. 2021.
\newblock \href {https://doi.org/10.18653/v1/2021.findings-emnlp.4} {Language
  clustering for multilingual named entity recognition}.
\newblock In \emph{Findings of the Association for Computational Linguistics:
  EMNLP 2021}, pages 40--45, Punta Cana, Dominican Republic. Association for
  Computational Linguistics.

\bibitem[{Shen et~al.(2020)Shen, Dong, Ye, Ma, Yao, Gholami, Mahoney, and
  Keutzer}]{shen2020q}
Sheng Shen, Zhen Dong, Jiayu Ye, Linjian Ma, Zhewei Yao, Amir Gholami,
  Michael~W Mahoney, and Kurt Keutzer. 2020.
\newblock Q-bert: Hessian based ultra low precision quantization of bert.
\newblock In \emph{Proceedings of the AAAI Conference on Artificial
  Intelligence}, volume~34, pages 8815--8821.

\bibitem[{Treviso et~al.(2022)Treviso, Ji, Lee, van Aken, Cao, Ciosici, Hassid,
  Heafield, Hooker, Martins, Martins, Milder, Raffel, Simpson, Slonim,
  Balasubramanian, Derczynski, and Schwartz}]{treviso2022}
Marcos Treviso, Tianchu Ji, Ji-Ung Lee, Betty van Aken, Qingqing Cao, Manuel~R.
  Ciosici, Michael Hassid, Kenneth Heafield, Sara Hooker, Pedro~H. Martins,
  André F.~T. Martins, Peter Milder, Colin Raffel, Edwin Simpson, Noam Slonim,
  Niranjan Balasubramanian, Leon Derczynski, and Roy Schwartz. 2022.
\newblock \href {https://doi.org/10.48550/ARXIV.2209.00099} {Efficient methods
  for natural language processing: A survey}.

\bibitem[{Tsai et~al.(2019)Tsai, Riesa, Johnson, Arivazhagan, Li, and
  Archer}]{tsai-etal-2019-small}
Henry Tsai, Jason Riesa, Melvin Johnson, Naveen Arivazhagan, Xin Li, and Amelia
  Archer. 2019.
\newblock \href {https://doi.org/10.18653/v1/D19-1374} {Small and practical
  {BERT} models for sequence labeling}.
\newblock In \emph{Proceedings of the 2019 Conference on Empirical Methods in
  Natural Language Processing and the 9th International Joint Conference on
  Natural Language Processing (EMNLP-IJCNLP)}, pages 3632--3636, Hong Kong,
  China. Association for Computational Linguistics.

\bibitem[{Wang et~al.(2020)Wang, K, Mayhew, and
  Roth}]{wang-etal-2020-extending}
Zihan Wang, Karthikeyan K, Stephen Mayhew, and Dan Roth. 2020.
\newblock \href {https://doi.org/10.18653/v1/2020.findings-emnlp.240}
  {Extending multilingual {BERT} to low-resource languages}.
\newblock In \emph{Findings of the Association for Computational Linguistics:
  EMNLP 2020}, pages 2649--2656, Online. Association for Computational
  Linguistics.

\bibitem[{Warden and Situnayake(2019)}]{warden2019tinyml}
P.~Warden and D.~Situnayake. 2019.
\newblock \href {https://books.google.com/books?id=sB3mxQEACAAJ} {\emph{TinyML:
  Machine Learning with TensorFlow Lite on Arduino and Ultra-Low-Power
  Microcontrollers}}.
\newblock O'Reilly Media, Incorporated.

\bibitem[{Wu and Dredze(2019)}]{wu-dredze-2019-beto}
Shijie Wu and Mark Dredze. 2019.
\newblock \href {https://doi.org/10.18653/v1/D19-1077} {Beto, bentz, becas: The
  surprising cross-lingual effectiveness of {BERT}}.
\newblock In \emph{Proceedings of the 2019 Conference on Empirical Methods in
  Natural Language Processing and the 9th International Joint Conference on
  Natural Language Processing (EMNLP-IJCNLP)}, pages 833--844, Hong Kong,
  China. Association for Computational Linguistics.

\bibitem[{Wu and Dredze(2020)}]{wu-dredze-2020-languages}
Shijie Wu and Mark Dredze. 2020.
\newblock \href {https://doi.org/10.18653/v1/2020.repl4nlp-1.16} {Are all
  languages created equal in multilingual {BERT}?}
\newblock In \emph{Proceedings of the 5th Workshop on Representation Learning
  for NLP}, pages 120--130, Online. Association for Computational Linguistics.

\bibitem[{Xu et~al.(2021)Xu, Zhou, Ge, Xu, McAuley, and
  Wei}]{xu-etal-2021-beyond}
Canwen Xu, Wangchunshu Zhou, Tao Ge, Ke~Xu, Julian McAuley, and Furu Wei. 2021.
\newblock \href {https://doi.org/10.18653/v1/2021.emnlp-main.832} {Beyond
  preserved accuracy: Evaluating loyalty and robustness of {BERT} compression}.
\newblock In \emph{Proceedings of the 2021 Conference on Empirical Methods in
  Natural Language Processing}, pages 10653--10659, Online and Punta Cana,
  Dominican Republic. Association for Computational Linguistics.

\bibitem[{Yaseen and Langer(2021)}]{Yaseen2021DataAF}
Usama Yaseen and Stefan Langer. 2021.
\newblock Data augmentation for low-resource named entity recognition using
  backtranslation.
\newblock \emph{ArXiv}, abs/2108.11703.

\bibitem[{Zhang et~al.(2022)Zhang, Roller, Goyal, Artetxe, Chen, Chen, Dewan,
  Diab, Li, Lin, Mihaylov, Ott, Shleifer, Shuster, Simig, Koura, Sridhar, Wang,
  and Zettlemoyer}]{Zhang2022OPTOP}
Susan Zhang, Stephen Roller, Naman Goyal, Mikel Artetxe, Moya Chen, Shuohui
  Chen, Christopher Dewan, Mona Diab, Xian Li, Xi~Victoria Lin, Todor Mihaylov,
  Myle Ott, Sam Shleifer, Kurt Shuster, Daniel Simig, Punit~Singh Koura, Anjali
  Sridhar, Tianlu Wang, and Luke Zettlemoyer. 2022.
\newblock Opt: Open pre-trained transformer language models.
\newblock \emph{ArXiv}, abs/2205.01068.

\end{thebibliography}
\bibliographystyle{acl_natbib}

\clearpage
\appendix

\section{Hyperparameters}\label{app:hyperparameters}

\subsection{Fine-tuning Hyperparameters}\label{app:ft_hyper}

\textbf{Train epochs}: 60 \\
\textbf{Optimizer}: AdamW~\cite{Loshchilov2019DecoupledWD} \\
\textbf{Learning rate}: 7e-5 \\
\textbf{Max sequence length}: 512 \\
\textbf{Dropout}: 0.1 \\
\textbf{Batch size}: \\
\indent Data size $\in \{100, 1000\}$: 8 \\
\indent Data size $\in \{5000\}$: 16 \\
\indent Data size $\in \{10000, 15000, 20000\}$: 16 \\

\subsection{Pruning Hyperparameters}\label{app:prune_hyper}
\textbf{Data size} = 100:\\
\indent pruning start step: 10 \\
\indent pruning end step: 60\\
\indent pruning frequency: 10 \\

\textbf{Data size} = 1000:\\
\indent pruning start step: 100 \\
\indent pruning end step: 300 \\
\indent pruning frequency: 50 \\

\textbf{Data size $\in \{5000, 10000\}$}:\\
\indent pruning start step: 500 \\
\indent pruning end step: 1200 \\
\indent pruning frequency: 100 \\

\textbf{Data size} = 15000:\\
\indent pruning start step: 700 \\
\indent pruning end step: 1800 \\
\indent pruning frequency: 150 \\

\textbf{Data size} = 20000:\\
\indent pruning start step: 1000 \\
\indent pruning end step: 2400 \\
\indent pruning frequency: 200 \\

\section{Additional Diagrams} \label{app:diagrams}

\begin{comment}
\paragraph{Relation of data size and sparsity}
Figure~\ref{fig:size_vs_f1} relates fine-tuning size and F1 score for a selection of dense and sparse models for all languages individually. It shows that the correlation between fine-tuning and F1 decreases with increasing sparsity.

\begin{figure}[ht]
    \centering
    \includegraphics[width=\columnwidth]{plots/ft_size_vs_f1.pdf}
    \caption{Fine-tuning Size vs. F1 for 0\%, 50\% and 90\% sparse models.}
    \label{fig:size_vs_f1}
\end{figure}

\begin{figure}[h!]
    \centering
    \includegraphics[width=\columnwidth]{plots/f1_embedding_pruning.pdf}
    \caption{Mean absolute difference in F1 of including embeddings in pruning versus excluding them. Languages are grouped according to their fine-tuning size. The shaded areas represent the standard deviation.}
    \label{fig:pruning_embeddings}
\end{figure}

\paragraph{Impact of pruning embeddings} 
Figure~\ref{fig:pruning_embeddings} depicts the relative change in F1 for pruning embeddings compared to not pruning them. We can see that higher-sparsity leads to a larger drop in performance when pruning embeddings.
\end{comment}

\begin{figure*}[ht!]
    \centering
    \includegraphics[width=0.9\textwidth]{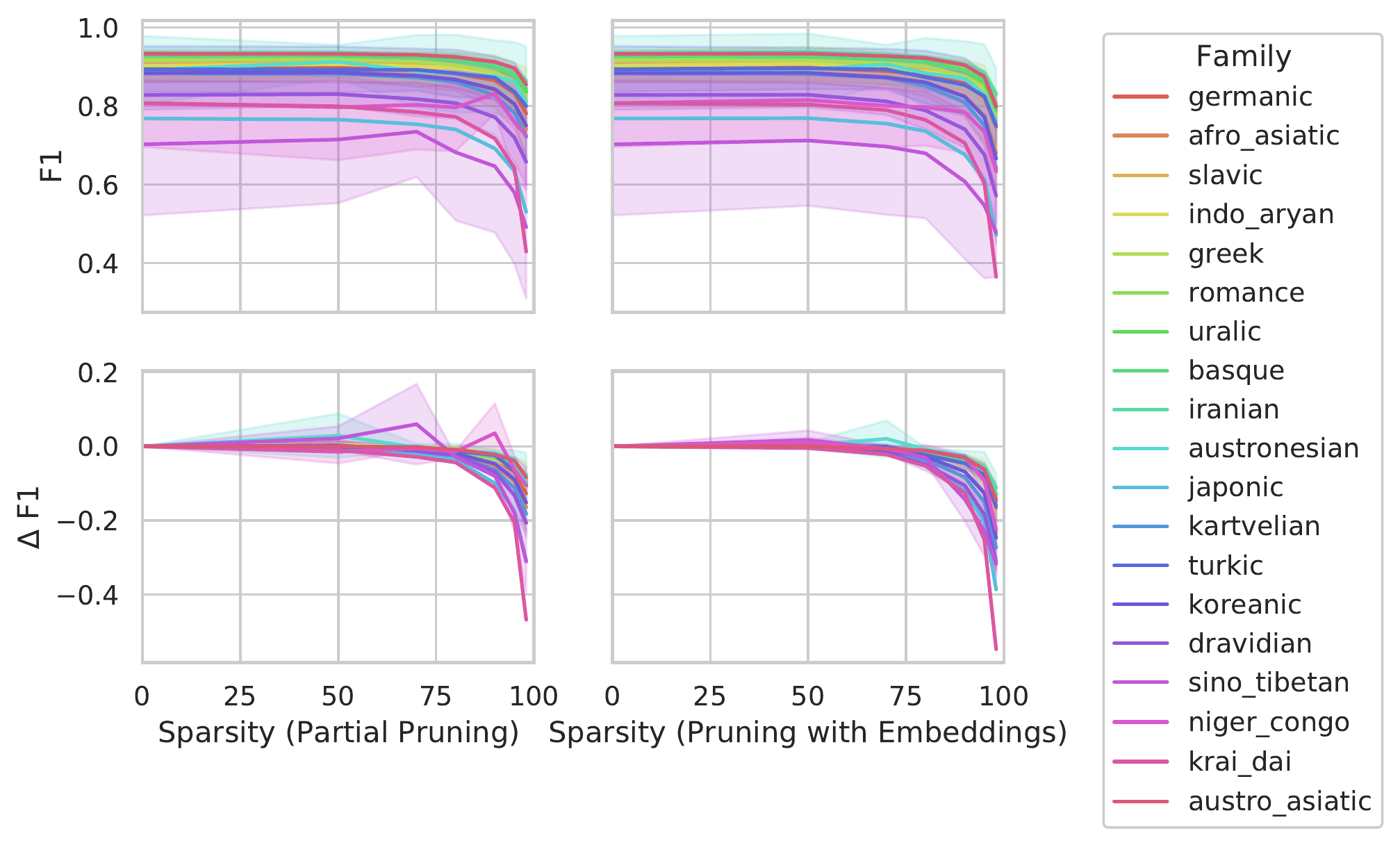}
    \caption{\textbf{Regular test:} Absolute F1 scores on top, relative differences in comparison to the dense model on the bottom. Results are averaged for languages grouped according to their \emph{language families}. The shaded areas represent the standard deviation.}
    \label{fig:comp_rel_pruning_family}
\end{figure*}

\begin{figure*}[h!]
    \centering
    \includegraphics[width=0.9\textwidth]{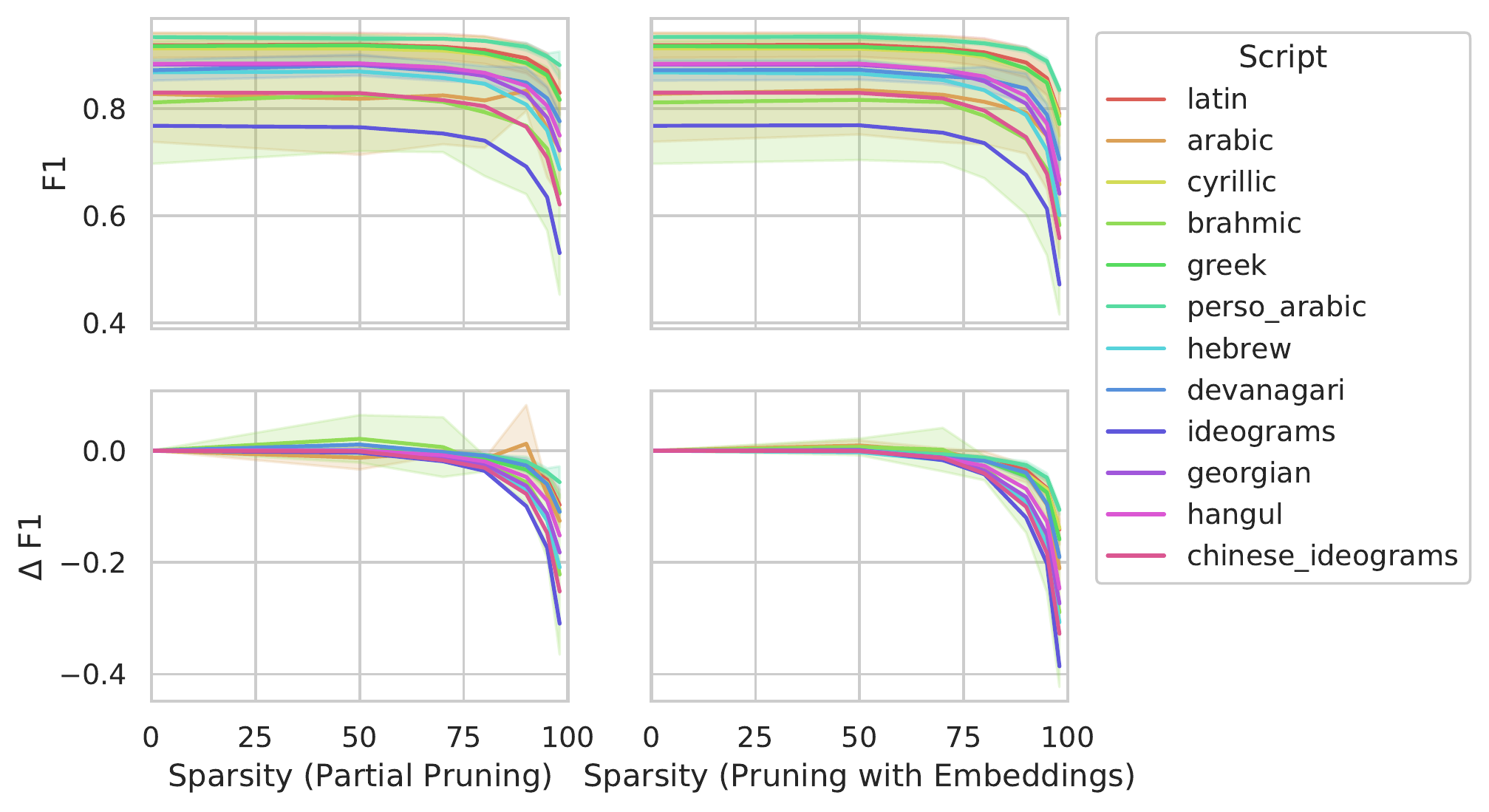}
    \caption{\textbf{Regular test:} Absolute F1 scores on top, relative differences in comparison to the dense model on the bottom. Results are averaged for languages grouped according to their \emph{script}. The shaded areas represent the standard deviation.}
    \label{fig:comp_rel_pruning_script}
\end{figure*}

\begin{figure*}[ht!]
    \centering
    \includegraphics[width=0.9\textwidth]{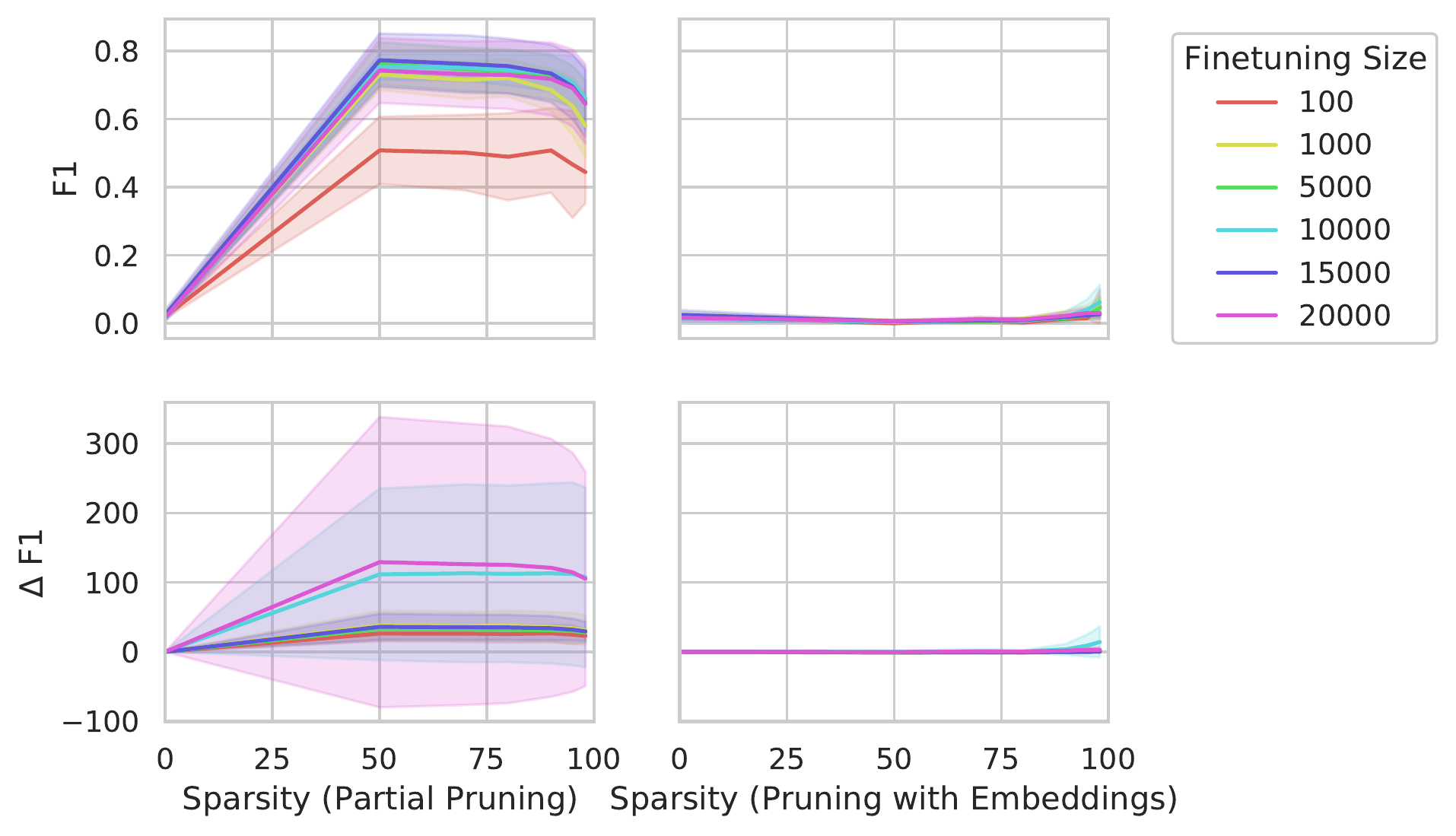}
    \caption{\textbf{In-language perturbation test:} Absolute F1 scores on top, relative differences in comparison to the dense model on the bottom. Results are averaged for languages grouped according to their \emph{fine-tuning size}. The shaded areas represent the standard deviation.}
    \label{fig:comp_rel_pruning_robust}
\end{figure*}

\begin{figure*}[ht!]
    \centering
    \includegraphics[width=0.9\textwidth]{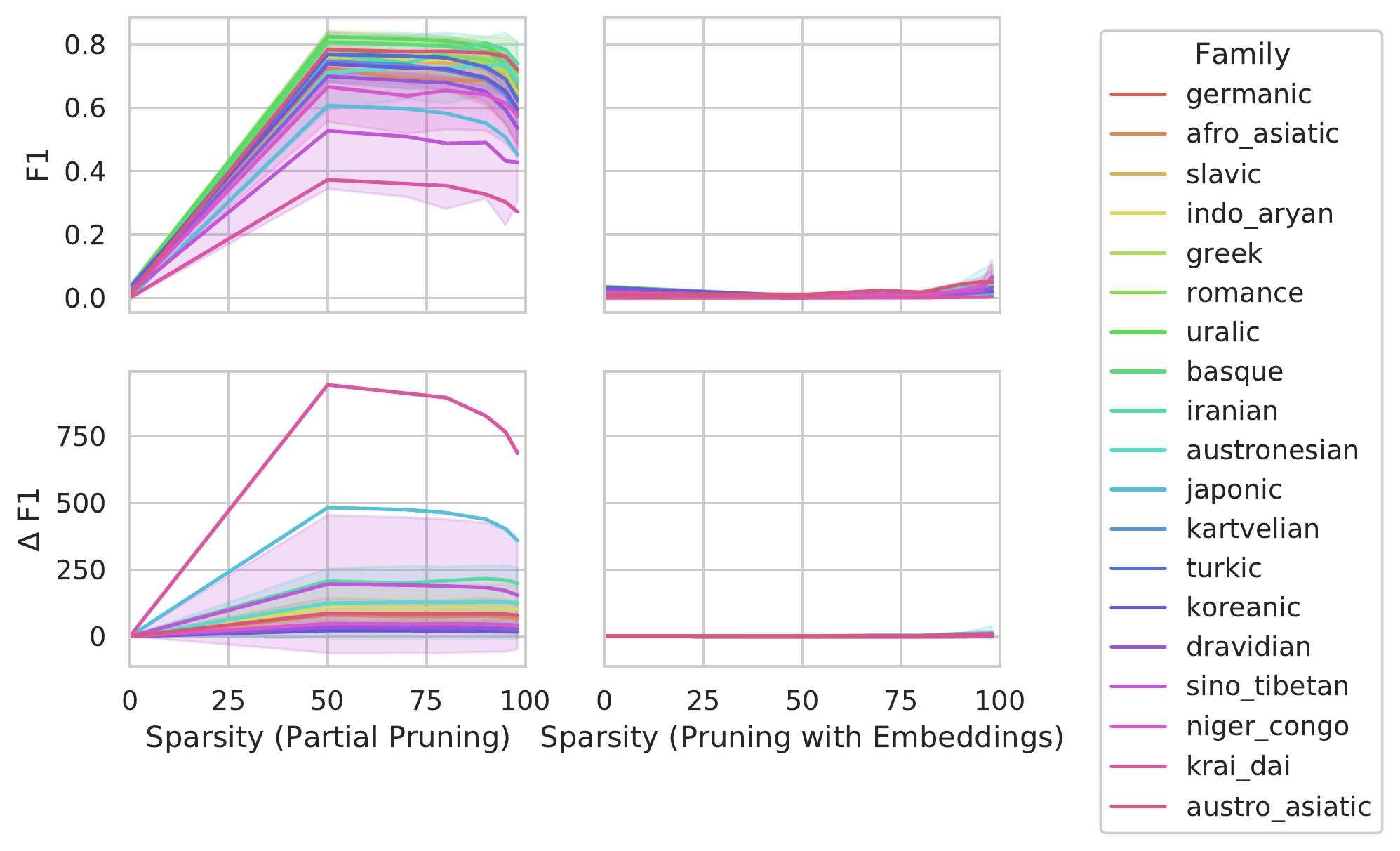}
    \caption{\textbf{In-language perturbation test:} Absolute F1 scores on top, relative differences in comparison to the dense model on the bottom. Results are averaged for languages grouped according to their \emph{language families}. The shaded areas represent the standard deviation.}
    \label{fig:comp_rel_pruning_family_robust}
\end{figure*}

\begin{figure*}[h!]
    \centering
    \includegraphics[width=0.9\textwidth]{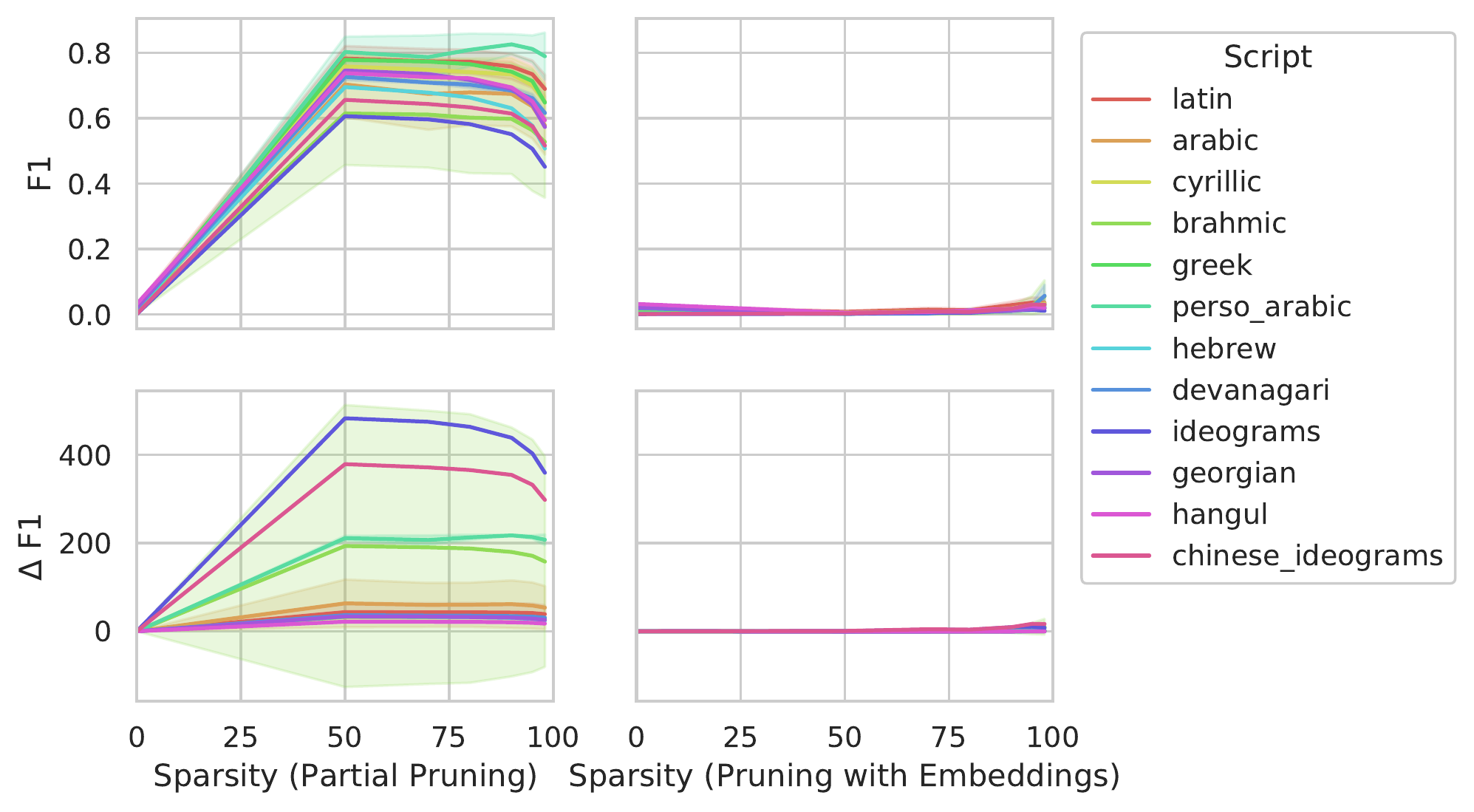}
    \caption{\textbf{In-language perturbation test:} Absolute F1 scores on top, relative differences in comparison to the dense model on the bottom. Results are averaged for languages grouped according to their \emph{script}. The shaded areas represent the standard deviation.}
    \label{fig:comp_rel_pruning_script_robust}
\end{figure*}

\begin{figure*}
    \centering
    \includegraphics[width=\textwidth]{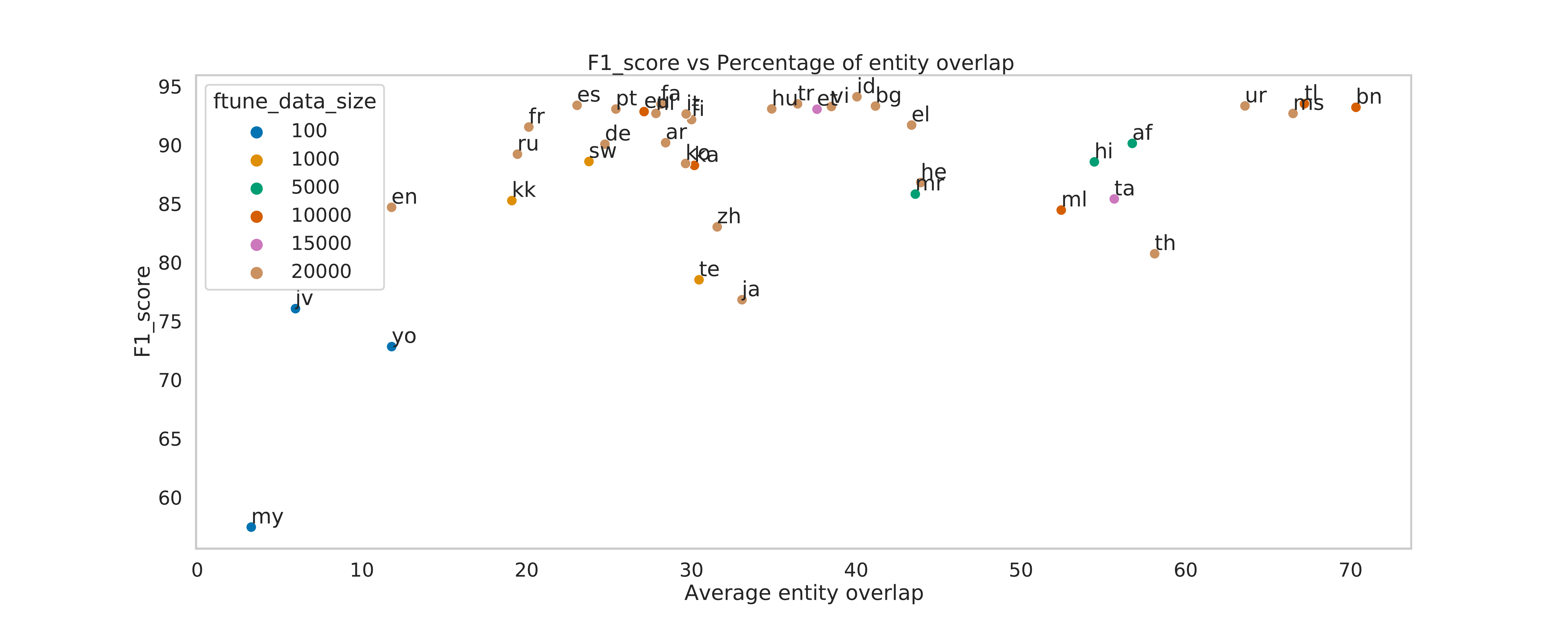}
    \caption{\textbf{Entity overlap:} Absolute F1 scores of dense multilingual model \textit{vs} percentage overlap of entities between train and test set. The colors indicate the size of finetuning data per language.}
    \label{fig:entity_overlap}
\end{figure*}

\paragraph{Relative change for different groups of languages}
Figures~\ref{fig:comp_rel_pruning_family} and ~\ref{fig:comp_rel_pruning_script} show the relative change in F1 compared to the dense model averaged across languages within the same family or with the same script, respectively, on the regular test set. Figures~\ref{fig:comp_rel_pruning_robust}, ~\ref{fig:comp_rel_pruning_family_robust} and ~\ref{fig:comp_rel_pruning_script_robust} depict the corresponding results on the in-language perturbed test sets. Figure ~\ref{fig:entity_overlap} shows the correlation between percentage entity overlap and F1 on dense multilingual models.

\section{Full Results}
% \julia{robustness and monolingual results need to be added}
\label{app:results}
We present the results for individual languages on both the regular and perturbed test sets obtained via multilingual finetuning in tables ~\ref{tab:multilingual_regular_without_embeddings}, ~\ref{tab:multilingual_regular_with_embeddings}, ~\ref{tab:multilingual_perturbed_without_embeddings} and ~\ref{tab:multilingual_perturbed_with_embeddings} \\

We present the results for individual languages on both the regular and perturbed test sets obtained via monolingual finetuning in tables ~\ref{tab:mono_regular_without_embeddings}, ~\ref{tab:mono_regular_with_embeddings}, ~\ref{tab:mono_perturbed_without_embeddings} and ~\ref{tab:mono_perturbed_with_embeddings} \\

\section{Examples of Perturbed Test Sentences}
\label{app:examples}
We present examples of perturbed test sentences in the \textit{in-language} setting for English (table \ref{tab:sample_examples_english}) and Yoruba language table (\ref{tab:sample_examples_yoruba}). 

\clearpage

\begin{table*}
\centering
\begin{tabular}{lrrrrrrr}
\toprule
languages &       0 &      50 &      70 &      80 &      90 &      95 &      98 \\
\midrule
       af &  0.9014 &  0.9164 &  0.9002 &  0.8944 &  0.8874 &  0.8571 &  0.8204 \\
       ar &  0.9020 &  0.9033 &  0.8983 &  0.8891 &  0.8719 &  0.8447 &  0.7943 \\
       bg &  0.9332 &  0.9317 &  0.9287 &  0.9253 &  0.9090 &  0.8881 &  0.8524 \\
       bn &  0.9321 &  0.9360 &  0.9326 &  0.9313 &  0.9191 &  0.9032 &  0.8760 \\
       de &  0.9007 &  0.9021 &  0.8968 &  0.8886 &  0.8683 &  0.8375 &  0.7823 \\
       el &  0.9170 &  0.9184 &  0.9133 &  0.9039 &  0.8856 &  0.8625 &  0.8164 \\
       en &  0.8470 &  0.8481 &  0.8404 &  0.8305 &  0.8036 &  0.7654 &  0.6951 \\
       es &  0.9338 &  0.9346 &  0.9295 &  0.9241 &  0.9139 &  0.8952 &  0.8612 \\
       et &  0.9305 &  0.9311 &  0.9283 &  0.9212 &  0.9057 &  0.8844 &  0.8433 \\
       eu &  0.9285 &  0.9271 &  0.9232 &  0.9165 &  0.9016 &  0.8788 &  0.8451 \\
       fa &  0.9348 &  0.9356 &  0.9305 &  0.9270 &  0.9107 &  0.8938 &  0.8634 \\
       fi &  0.9217 &  0.9210 &  0.9178 &  0.9103 &  0.8918 &  0.8666 &  0.8220 \\
       fr &  0.9153 &  0.9154 &  0.9109 &  0.9069 &  0.8892 &  0.8618 &  0.8168 \\
       he &  0.8681 &  0.8696 &  0.8578 &  0.8466 &  0.8078 &  0.7596 &  0.6866 \\
       hi &  0.8857 &  0.8955 &  0.8823 &  0.8748 &  0.8692 &  0.8348 &  0.7880 \\
       hu &  0.9308 &  0.9331 &  0.9290 &  0.9214 &  0.9057 &  0.8803 &  0.8396 \\
       id &  0.9411 &  0.9401 &  0.9385 &  0.9335 &  0.9247 &  0.9094 &  0.8809 \\
       it &  0.9265 &  0.9261 &  0.9224 &  0.9146 &  0.8992 &  0.8721 &  0.8280 \\
       ja &  0.7683 &  0.7655 &  0.7537 &  0.7404 &  0.6918 &  0.6346 &  0.5303 \\
       jv &  0.7607 &  0.8505 &  0.7509 &  0.7345 &  0.7375 &  0.7138 &  0.5987 \\
       ka &  0.8827 &  0.8823 &  0.8727 &  0.8612 &  0.8270 &  0.7833 &  0.7216 \\
       kk &  0.8527 &  0.8510 &  0.8544 &  0.8404 &  0.8343 &  0.7834 &  0.7522 \\
       ko &  0.8843 &  0.8848 &  0.8771 &  0.8672 &  0.8428 &  0.8053 &  0.7499 \\
       ml &  0.8446 &  0.8462 &  0.8365 &  0.8281 &  0.7972 &  0.7505 &  0.6805 \\
       mr &  0.8582 &  0.8676 &  0.8572 &  0.8530 &  0.8284 &  0.8043 &  0.7650 \\
       ms &  0.9269 &  0.9219 &  0.9361 &  0.9336 &  0.9099 &  0.8965 &  0.8679 \\
       my &  0.5746 &  0.6003 &  0.6532 &  0.5594 &  0.5274 &  0.4516 &  0.3622 \\
       nl &  0.9269 &  0.9264 &  0.9233 &  0.9188 &  0.9013 &  0.8709 &  0.8257 \\
       pt &  0.9306 &  0.9335 &  0.9292 &  0.9252 &  0.9118 &  0.8918 &  0.8516 \\
       ru &  0.8922 &  0.8930 &  0.8890 &  0.8770 &  0.8598 &  0.8317 &  0.7823 \\
       sw &  0.8860 &  0.8924 &  0.8837 &  0.8751 &  0.8671 &  0.8530 &  0.8231 \\
       ta &  0.8541 &  0.8486 &  0.8484 &  0.8319 &  0.7984 &  0.7607 &  0.7019 \\
       te &  0.7853 &  0.7958 &  0.7678 &  0.7621 &  0.7192 &  0.6483 &  0.5907 \\
       th &  0.8074 &  0.7993 &  0.7845 &  0.7724 &  0.7171 &  0.6424 &  0.4293 \\
       tl &  0.9352 &  0.9389 &  0.9292 &  0.9289 &  0.9287 &  0.9300 &  0.8946 \\
       tr &  0.9351 &  0.9338 &  0.9301 &  0.9256 &  0.9105 &  0.8887 &  0.8478 \\
       ur &  0.9333 &  0.9269 &  0.9310 &  0.9266 &  0.9208 &  0.9018 &  0.8994 \\
       vi &  0.9328 &  0.9326 &  0.9302 &  0.9247 &  0.9123 &  0.8966 &  0.8549 \\
       yo &  0.7284 &  0.7015 &  0.7225 &  0.7172 &  0.7956 &  0.6635 &  0.6264 \\
       zh &  0.8303 &  0.8293 &  0.8162 &  0.8048 &  0.7661 &  0.7080 &  0.6209 \\
      \midrule
    means &  0.8795 &  0.8827 &  0.8764 &  0.8667 &  0.8492 &  0.8151 &  0.7622 \\
  medians &  0.9017 &  0.9093 &  0.8993 &  0.8918 &  0.8788 &  0.8551 &  0.8166 \\
\bottomrule
\end{tabular}

\caption{F1 scores for multilingual fine-tuning on the regular data for various levels of sparsity without pruning embedding layers.}
\label{tab:multilingual_regular_without_embeddings}
\end{table*}

\begin{table*}
\centering

\begin{tabular}{lrrrrrrr}
\toprule
languages &       0 &      50 &      70 &      80 &      90 &      95 &      98 \\
\midrule
       af &  0.9014 &  0.9134 &  0.8960 &  0.8870 &  0.8810 &  0.8412 &  0.7878 \\
       ar &  0.9020 &  0.9034 &  0.8955 &  0.8849 &  0.8624 &  0.8279 &  0.7593 \\
       bg &  0.9332 &  0.9320 &  0.9270 &  0.9196 &  0.9018 &  0.8777 &  0.8222 \\
       bn &  0.9321 &  0.9543 &  0.9359 &  0.9197 &  0.9029 &  0.8951 &  0.8078 \\
       de &  0.9007 &  0.9006 &  0.8959 &  0.8854 &  0.8547 &  0.8227 &  0.7377 \\
       el &  0.9170 &  0.9158 &  0.9089 &  0.9006 &  0.8752 &  0.8483 &  0.7714 \\
       en &  0.8470 &  0.8491 &  0.8415 &  0.8283 &  0.7988 &  0.7604 &  0.6677 \\
       es &  0.9338 &  0.9316 &  0.9275 &  0.9236 &  0.9078 &  0.8896 &  0.8377 \\
       et &  0.9305 &  0.9305 &  0.9244 &  0.9172 &  0.8926 &  0.8642 &  0.7946 \\
       eu &  0.9285 &  0.9260 &  0.9193 &  0.9128 &  0.8903 &  0.8640 &  0.8059 \\
       fa &  0.9348 &  0.9379 &  0.9307 &  0.9244 &  0.9064 &  0.8843 &  0.8301 \\
       fi &  0.9217 &  0.9202 &  0.9139 &  0.9067 &  0.8814 &  0.8505 &  0.7814 \\
       fr &  0.9153 &  0.9147 &  0.9090 &  0.8983 &  0.8805 &  0.8525 &  0.7869 \\
       he &  0.8681 &  0.8656 &  0.8537 &  0.8346 &  0.7880 &  0.7226 &  0.6012 \\
       hi &  0.8857 &  0.8858 &  0.8709 &  0.8718 &  0.8578 &  0.8028 &  0.7212 \\
       hu &  0.9308 &  0.9302 &  0.9257 &  0.9189 &  0.8943 &  0.8628 &  0.7952 \\
       id &  0.9411 &  0.9400 &  0.9385 &  0.9342 &  0.9204 &  0.9014 &  0.8521 \\
       it &  0.9265 &  0.9253 &  0.9214 &  0.9136 &  0.8941 &  0.8602 &  0.7886 \\
       ja &  0.7683 &  0.7691 &  0.7552 &  0.7357 &  0.6761 &  0.6129 &  0.4716 \\
       jv &  0.7607 &  0.7576 &  0.8329 &  0.7503 &  0.7273 &  0.6433 &  0.5623 \\
       ka &  0.8827 &  0.8821 &  0.8718 &  0.8511 &  0.8096 &  0.7502 &  0.6412 \\
       kk &  0.8527 &  0.8585 &  0.8567 &  0.8258 &  0.8053 &  0.7821 &  0.7140 \\
       ko &  0.8843 &  0.8844 &  0.8727 &  0.8602 &  0.8238 &  0.7720 &  0.6660 \\
       ml &  0.8446 &  0.8425 &  0.8261 &  0.8172 &  0.7695 &  0.7139 &  0.6184 \\
       mr &  0.8582 &  0.8597 &  0.8504 &  0.8406 &  0.8178 &  0.7745 &  0.6905 \\
       ms &  0.9269 &  0.9402 &  0.9198 &  0.9200 &  0.9091 &  0.8757 &  0.8229 \\
       my &  0.5746 &  0.5948 &  0.5741 &  0.5627 &  0.4686 &  0.4160 &  0.3978 \\
       nl &  0.9269 &  0.9266 &  0.9226 &  0.9151 &  0.8949 &  0.8648 &  0.7951 \\
       pt &  0.9306 &  0.9318 &  0.9273 &  0.9216 &  0.9069 &  0.8831 &  0.8182 \\
       ru &  0.8922 &  0.8923 &  0.8854 &  0.8750 &  0.8489 &  0.8220 &  0.7504 \\
       sw &  0.8860 &  0.8880 &  0.8753 &  0.8659 &  0.8571 &  0.8332 &  0.7648 \\
       ta &  0.8541 &  0.8512 &  0.8365 &  0.8161 &  0.7662 &  0.7078 &  0.6072 \\
       te &  0.7853 &  0.7923 &  0.7725 &  0.7326 &  0.6867 &  0.6071 &  0.4890 \\
       th &  0.8074 &  0.8039 &  0.7896 &  0.7646 &  0.7059 &  0.6036 &  0.3651 \\
       tl &  0.9352 &  0.9360 &  0.9324 &  0.9310 &  0.9217 &  0.9041 &  0.8123 \\
       tr &  0.9351 &  0.9330 &  0.9301 &  0.9208 &  0.9017 &  0.8677 &  0.7862 \\
       ur &  0.9333 &  0.9313 &  0.9256 &  0.9200 &  0.9137 &  0.8934 &  0.8398 \\
       vi &  0.9328 &  0.9334 &  0.9270 &  0.9222 &  0.9042 &  0.8745 &  0.7975 \\
       yo &  0.7284 &  0.7426 &  0.7261 &  0.7279 &  0.7109 &  0.6368 &  0.5016 \\
       zh &  0.8303 &  0.8296 &  0.8194 &  0.7964 &  0.7469 &  0.6782 &  0.5581 \\
       \midrule
    means &  0.8795 &  0.8814 &  0.8741 &  0.8614 &  0.8341 &  0.7936 &  0.7105 \\
  medians &  0.9017 &  0.9084 &  0.8960 &  0.8862 &  0.8688 &  0.8372 &  0.7681 \\
\bottomrule
\end{tabular}

\caption{F1 scores for multilingual fine-tuning on the regular data for various levels of sparsity with pruning embedding layers.}
\label{tab:multilingual_regular_with_embeddings}

\end{table*}

\begin{table*}
\centering

\begin{tabular}{lrrrrrrr}
\toprule
languages &       0 &      50 &      70 &      80 &      90 &      95 &      98 \\
\midrule
       af &  0.0314 &  0.8349 &  0.8193 &  0.8142 &  0.7988 &  0.7648 &  0.7240 \\
       ar &  0.0060 &  0.7543 &  0.7046 &  0.7091 &  0.7426 &  0.7133 &  0.6623 \\
       bg &  0.0237 &  0.7829 &  0.7712 &  0.7711 &  0.7702 &  0.7400 &  0.6911 \\
       bn &  0.0055 &  0.7619 &  0.7489 &  0.7568 &  0.7620 &  0.7641 &  0.7289 \\
       de &  0.0257 &  0.8019 &  0.7946 &  0.7849 &  0.7562 &  0.7187 &  0.6690 \\
       el &  0.0230 &  0.7792 &  0.7737 &  0.7659 &  0.7429 &  0.7139 &  0.6481 \\
       en &  0.0143 &  0.6843 &  0.6781 &  0.6645 &  0.6407 &  0.6128 &  0.5552 \\
       es &  0.0119 &  0.7803 &  0.7666 &  0.7767 &  0.7790 &  0.7630 &  0.7207 \\
       et &  0.0350 &  0.8283 &  0.8216 &  0.8125 &  0.7939 &  0.7635 &  0.7161 \\
       eu &  0.0207 &  0.8065 &  0.7996 &  0.7945 &  0.7773 &  0.7403 &  0.6806 \\
       fa &  0.0037 &  0.7696 &  0.7405 &  0.7744 &  0.8042 &  0.7827 &  0.7385 \\
       fi &  0.0390 &  0.8398 &  0.8337 &  0.8264 &  0.8070 &  0.7722 &  0.7246 \\
       fr &  0.0221 &  0.7632 &  0.7551 &  0.7528 &  0.7405 &  0.7157 &  0.6780 \\
       he &  0.0201 &  0.6957 &  0.6788 &  0.6638 &  0.6310 &  0.5774 &  0.5076 \\
       hi &  0.0196 &  0.7199 &  0.7073 &  0.7102 &  0.6765 &  0.6526 &  0.6200 \\
       hu &  0.0316 &  0.8044 &  0.7952 &  0.7933 &  0.7793 &  0.7471 &  0.6948 \\
       id &  0.0118 &  0.8038 &  0.7921 &  0.7979 &  0.7916 &  0.7801 &  0.7375 \\
       it &  0.0226 &  0.7867 &  0.7756 &  0.7723 &  0.7511 &  0.7259 &  0.6807 \\
       ja &  0.0013 &  0.6068 &  0.5967 &  0.5824 &  0.5513 &  0.5067 &  0.4518 \\
       jv &  0.0161 &  0.5384 &  0.5771 &  0.5588 &  0.5972 &  0.5862 &  0.5102 \\
       ka &  0.0216 &  0.7465 &  0.7356 &  0.7174 &  0.6901 &  0.6415 &  0.5734 \\
       kk &  0.0242 &  0.7693 &  0.7667 &  0.7620 &  0.7208 &  0.6703 &  0.5889 \\
       ko &  0.0324 &  0.7384 &  0.7259 &  0.7227 &  0.6946 &  0.6520 &  0.5940 \\
       ml &  0.0215 &  0.6995 &  0.6962 &  0.6806 &  0.6653 &  0.6103 &  0.5482 \\
       mr &  0.0192 &  0.7342 &  0.7113 &  0.6959 &  0.6931 &  0.6709 &  0.6129 \\
       ms &  0.0094 &  0.7493 &  0.7642 &  0.7757 &  0.7597 &  0.7902 &  0.7403 \\
       my &  0.0276 &  0.3975 &  0.3742 &  0.3414 &  0.3658 &  0.2884 &  0.3389 \\
       nl &  0.0233 &  0.7759 &  0.7663 &  0.7628 &  0.7503 &  0.7149 &  0.6662 \\
       pt &  0.0170 &  0.7586 &  0.7453 &  0.7452 &  0.7394 &  0.7138 &  0.6908 \\
       ru &  0.0188 &  0.7349 &  0.7264 &  0.7116 &  0.6993 &  0.6621 &  0.6095 \\
       sw &  0.0118 &  0.7434 &  0.7217 &  0.7415 &  0.7210 &  0.7015 &  0.6716 \\
       ta &  0.0142 &  0.7174 &  0.7021 &  0.6987 &  0.6740 &  0.6276 &  0.5759 \\
       te &  0.0304 &  0.6803 &  0.6564 &  0.6581 &  0.6143 &  0.5424 &  0.4819 \\
       th &  0.0004 &  0.3727 &  0.3600 &  0.3537 &  0.3266 &  0.3028 &  0.2716 \\
       tl &  0.0024 &  0.7526 &  0.7777 &  0.7707 &  0.7826 &  0.7873 &  0.7679 \\
       tr &  0.0254 &  0.7667 &  0.7596 &  0.7530 &  0.7354 &  0.7095 &  0.6588 \\
       ur &  0.0039 &  0.8362 &  0.8343 &  0.8449 &  0.8486 &  0.8417 &  0.8412 \\
       vi &  0.0090 &  0.7831 &  0.7768 &  0.7779 &  0.7734 &  0.7612 &  0.7208 \\
       yo &  0.0172 &  0.5882 &  0.5532 &  0.5675 &  0.5609 &  0.5259 &  0.4841 \\
       zh &  0.0017 &  0.6567 &  0.6440 &  0.6336 &  0.6146 &  0.5758 &  0.5165 \\
     \midrule
    means &  0.0179 &  0.7286 &  0.7182 &  0.7149 &  0.7031 &  0.6733 &  0.6273 \\
  medians &  0.0194 &  0.7564 &  0.7471 &  0.7529 &  0.7399 &  0.7135 &  0.6642 \\
\bottomrule
\end{tabular}

\caption{F1 scores for multilingual fine-tuning on the perturbed data for various levels of sparsity without pruning embedding layers.}
\label{tab:multilingual_perturbed_without_embeddings}

\end{table*}

\begin{table*}
\centering

\begin{tabular}{lrrrrrrr}
\toprule
languages &       0 &      50 &      70 &      80 &      90 &      95 &      98 \\
\midrule
       af &  0.0314 &  0.0058 &  0.0076 &  0.0049 &  0.0243 &  0.0239 &  0.0228 \\
       ar &  0.0060 &  0.0058 &  0.0076 &  0.0104 &  0.0134 &  0.0168 &  0.0202 \\
       bg &  0.0237 &  0.0048 &  0.0070 &  0.0081 &  0.0129 &  0.0219 &  0.0239 \\
       bn &  0.0055 &  0.0008 &  0.0028 &  0.0013 &  0.0113 &  0.0494 &  0.1111 \\
       de &  0.0257 &  0.0055 &  0.0145 &  0.0113 &  0.0251 &  0.0305 &  0.0234 \\
       el &  0.0230 &  0.0035 &  0.0082 &  0.0094 &  0.0120 &  0.0152 &  0.0192 \\
       en &  0.0143 &  0.0082 &  0.0214 &  0.0128 &  0.0398 &  0.0490 &  0.0439 \\
       es &  0.0119 &  0.0069 &  0.0149 &  0.0146 &  0.0274 &  0.0395 &  0.0410 \\
       et &  0.0350 &  0.0072 &  0.0110 &  0.0123 &  0.0202 &  0.0251 &  0.0233 \\
       eu &  0.0207 &  0.0061 &  0.0106 &  0.0097 &  0.0224 &  0.0266 &  0.0303 \\
       fa &  0.0037 &  0.0038 &  0.0037 &  0.0075 &  0.0094 &  0.0280 &  0.0357 \\
       fi &  0.0390 &  0.0051 &  0.0112 &  0.0116 &  0.0190 &  0.0219 &  0.0213 \\
       fr &  0.0221 &  0.0102 &  0.0190 &  0.0131 &  0.0366 &  0.0457 &  0.0436 \\
       he &  0.0201 &  0.0029 &  0.0073 &  0.0067 &  0.0141 &  0.0212 &  0.0242 \\
       hi &  0.0196 &  0.0026 &  0.0024 &  0.0096 &  0.0155 &  0.0326 &  0.0815 \\
       hu &  0.0316 &  0.0058 &  0.0087 &  0.0118 &  0.0166 &  0.0182 &  0.0177 \\
       id &  0.0118 &  0.0112 &  0.0137 &  0.0082 &  0.0149 &  0.0247 &  0.0227 \\
       it &  0.0226 &  0.0098 &  0.0164 &  0.0147 &  0.0317 &  0.0352 &  0.0332 \\
       ja &  0.0013 &  0.0021 &  0.0059 &  0.0054 &  0.0130 &  0.0144 &  0.0115 \\
       jv &  0.0161 &  0.0000 &  0.0156 &  0.0000 &  0.0098 &  0.0162 &  0.0042 \\
       ka &  0.0216 &  0.0037 &  0.0073 &  0.0069 &  0.0119 &  0.0172 &  0.0190 \\
       kk &  0.0242 &  0.0061 &  0.0048 &  0.0148 &  0.0137 &  0.0184 &  0.0219 \\
       ko &  0.0324 &  0.0058 &  0.0075 &  0.0128 &  0.0178 &  0.0261 &  0.0210 \\
       ml &  0.0215 &  0.0014 &  0.0028 &  0.0034 &  0.0063 &  0.0176 &  0.0255 \\
       mr &  0.0192 &  0.0022 &  0.0033 &  0.0159 &  0.0066 &  0.0141 &  0.0332 \\
       ms &  0.0094 &  0.0178 &  0.0286 &  0.0295 &  0.0489 &  0.0738 &  0.0586 \\
       my &  0.0276 &  0.0000 &  0.0130 &  0.0078 &  0.0222 &  0.0104 &  0.1038 \\
       nl &  0.0233 &  0.0074 &  0.0157 &  0.0131 &  0.0284 &  0.0313 &  0.0291 \\
       pt &  0.0170 &  0.0106 &  0.0181 &  0.0158 &  0.0379 &  0.0515 &  0.0532 \\
       ru &  0.0188 &  0.0072 &  0.0122 &  0.0103 &  0.0249 &  0.0374 &  0.0440 \\
       sw &  0.0118 &  0.0112 &  0.0137 &  0.0141 &  0.0405 &  0.0555 &  0.0798 \\
       ta &  0.0142 &  0.0048 &  0.0060 &  0.0065 &  0.0171 &  0.0224 &  0.0308 \\
       te &  0.0304 &  0.0038 &  0.0083 &  0.0117 &  0.0154 &  0.0208 &  0.0437 \\
       th &  0.0004 &  0.0003 &  0.0010 &  0.0009 &  0.0025 &  0.0026 &  0.0034 \\
       tl &  0.0024 &  0.0075 &  0.0179 &  0.0118 &  0.0437 &  0.0892 &  0.1235 \\
       tr &  0.0254 &  0.0043 &  0.0063 &  0.0090 &  0.0143 &  0.0174 &  0.0167 \\
       ur &  0.0039 &  0.0018 &  0.0047 &  0.0028 &  0.0137 &  0.0269 &  0.0246 \\
       vi &  0.0090 &  0.0095 &  0.0234 &  0.0175 &  0.0424 &  0.0504 &  0.0491 \\
       yo &  0.0172 &  0.0000 &  0.0000 &  0.0000 &  0.0083 &  0.0187 &  0.0401 \\
       zh &  0.0017 &  0.0031 &  0.0098 &  0.0083 &  0.0179 &  0.0309 &  0.0304 \\
     \midrule
    means &  0.0179 &  0.0054 &  0.0103 &  0.0099 &  0.0206 &  0.0297 &  0.0377 \\
  medians &  0.0194 &  0.0053 &  0.0085 &  0.0100 &  0.0168 &  0.0249 &  0.0297 \\
\bottomrule
\end{tabular}

\caption{F1 scores for multilingual fine-tuning on the perturbed data for various levels of sparsity with pruning embedding layers.}
\label{tab:multilingual_perturbed_with_embeddings}

\end{table*}

\begin{table*}
\centering
\begin{tabular}{lrrrrrrr}
\toprule
languages &       0 &      50 &      70 &      80 &      90 &      95 &      98 \\
\midrule
       en &  0.8468 &  0.8421 &  0.8283 &  0.7987 &  0.7049 &  0.5618 &  0.5592 \\
       zh &  0.8299 &  0.8262 &  0.8057 &  0.7726 &  0.6490 &  0.4759 &  0.4159 \\
       bn &  0.9284 &  0.9319 &  0.9205 &  0.9130 &  0.8619 &  0.7773 &  0.6028 \\
       eu &  0.9236 &  0.9179 &  0.9084 &  0.8904 &  0.8264 &  0.7209 &  0.6641 \\
       af &  0.9044 &  0.8970 &  0.8927 &  0.8878 &  0.7944 &  0.6800 &  0.6740 \\
       hi &  0.8827 &  0.9083 &  0.8643 &  0.8357 &  0.7579 &  0.6267 &  0.5863 \\
       sw &  0.8617 &  0.8541 &  0.8553 &  0.8496 &  0.7554 &  0.7017 &  0.6900 \\
       te &  0.7687 &  0.7481 &  0.7383 &  0.6859 &  0.4619 &  0.4864 &  0.4667 \\
       jv &  0.5478 &  0.5044 &  0.4976 &  0.3387 &  0.3210 &  0.3883 &  0.4025 \\
       yo &  0.7207 &  0.6266 &  0.6246 &  0.6387 &  0.5439 &  0.6567 &  0.5374 \\
\midrule
    means &  0.8215 &  0.8057 &  0.7936 &  0.7611 &  0.6677 &  0.6076 &  0.5599 \\
  medians &  0.8543 &  0.8481 &  0.8418 &  0.8172 &  0.7302 &  0.6417 &  0.5728 \\
\bottomrule
\end{tabular}

\caption{F1 scores for monolingual fine-tuning on the regular data for various levels of sparsity without pruning embedding layers.}
\label{tab:mono_regular_without_embeddings}
\end{table*}

\begin{table*}
\centering
\begin{tabular}{lrrrrrrr}
\toprule
languages &       0 &      50 &      70 &      80 &      90 &      95 &      98 \\
\midrule
       en &  0.8468 &  0.8382 &  0.8157 &  0.7828 &  0.6560 &  0.5730 &  0.5593 \\
       zh &  0.8299 &  0.8255 &  0.7974 &  0.7488 &  0.6073 &  0.4329 &  0.4075 \\
       bn &  0.9284 &  0.9421 &  0.9179 &  0.9033 &  0.8288 &  0.6957 &  0.5463 \\
       eu &  0.9236 &  0.9189 &  0.9051 &  0.8822 &  0.7885 &  0.6631 &  0.6583 \\
       af &  0.9044 &  0.8981 &  0.8807 &  0.8512 &  0.7464 &  0.6460 &  0.6632 \\
       hi &  0.8827 &  0.8670 &  0.8612 &  0.8224 &  0.7198 &  0.5636 &  0.5977 \\
       sw &  0.8617 &  0.8751 &  0.8508 &  0.8148 &  0.7158 &  0.6870 &  0.6980 \\
       te &  0.7687 &  0.7592 &  0.7188 &  0.6253 &  0.4601 &  0.4951 &  0.4857 \\
       jv &  0.5478 &  0.5123 &  0.5055 &  0.3758 &  0.3494 &  0.4369 &  0.2939 \\
       yo &  0.7207 &  0.6292 &  0.6271 &  0.6480 &  0.5786 &  0.5428 &  0.5779 \\
\midrule
    means &  0.8215 &  0.8066 &  0.7880 &  0.7454 &  0.6451 &  0.5736 &  0.5488 \\
  medians &  0.8543 &  0.8526 &  0.8332 &  0.7988 &  0.6859 &  0.5683 &  0.5686 \\
\bottomrule
\end{tabular}
\caption{F1 scores for monolingual fine-tuning on the regular data for various levels of sparsity with pruning embedding layers.}
\label{tab:mono_regular_with_embeddings}
\end{table*}

\begin{table*}
\centering
\begin{tabular}{lrrrrrrr}
\toprule
languages &       0 &      50 &      70 &      80 &      90 &      95 &      98 \\
\midrule
       en &  0.0230 &  0.7032 &  0.6841 &  0.6570 &  0.5543 &  0.4206 &  0.3337 \\
       zh &  0.0055 &  0.6551 &  0.6492 &  0.6245 &  0.5477 &  0.4262 &  0.2884 \\
       bn &  0.0138 &  0.8090 &  0.8000 &  0.7783 &  0.7106 &  0.6376 &  0.4981 \\
       eu &  0.0180 &  0.7938 &  0.7782 &  0.7470 &  0.6502 &  0.5274 &  0.3788 \\
       af &  0.0271 &  0.8260 &  0.8185 &  0.7960 &  0.6921 &  0.5562 &  0.4475 \\
       hi &  0.0166 &  0.7289 &  0.7094 &  0.6852 &  0.5889 &  0.4672 &  0.2965 \\
       sw &  0.0214 &  0.7326 &  0.7490 &  0.6785 &  0.5173 &  0.4733 &  0.3058 \\
       te &  0.0229 &  0.6581 &  0.6095 &  0.5602 &  0.3482 &  0.2932 &  0.1049 \\
       jv &  0.0223 &  0.4165 &  0.3449 &  0.2146 &  0.1439 &  0.0000 &  0.0000 \\
       yo &  0.0187 &  0.5396 &  0.5371 &  0.4288 &  0.3087 &  0.0168 &  0.0000 \\
\midrule
    means &  0.0189 &  0.6863 &  0.6680 &  0.6170 &  0.5062 &  0.3818 &  0.2654 \\
  medians &  0.0201 &  0.7160 &  0.6967 &  0.6678 &  0.5510 &  0.4467 &  0.3011 \\
\bottomrule
\end{tabular}
\caption{F1 scores for monolingual fine-tuning on the perturbed data for various levels of sparsity without pruning embedding layers.}
\label{tab:mono_perturbed_without_embeddings}
\end{table*}

\begin{table*}
\centering
\begin{tabular}{lrrrrrrr}
\toprule
languages &       0 &      50 &      70 &      80 &      90 &      95 &      98 \\
\midrule
       en &  0.0230 &  0.0634 &  0.0465 &  0.0418 &  0.0401 &  0.0351 &  0.0231 \\
       zh &  0.0055 &  0.0115 &  0.0160 &  0.0209 &  0.0316 &  0.0213 &  0.0101 \\
       bn &  0.0138 &  0.0000 &  0.0124 &  0.0124 &  0.0009 &  0.0055 &  0.0020 \\
       eu &  0.0180 &  0.0060 &  0.0116 &  0.0146 &  0.0202 &  0.0185 &  0.0249 \\
       af &  0.0271 &  0.0013 &  0.0084 &  0.0190 &  0.0132 &  0.0197 &  0.0262 \\
       hi &  0.0166 &  0.0337 &  0.0382 &  0.0104 &  0.0007 &  0.0041 &  0.0184 \\
       sw &  0.0214 &  0.0092 &  0.0541 &  0.0526 &  0.0469 &  0.0576 &  0.0210 \\
       te &  0.0229 &  0.0007 &  0.0029 &  0.0070 &  0.0016 &  0.0000 &  0.0000 \\
       jv &  0.0223 &  0.0212 &  0.0074 &  0.0114 &  0.0034 &  0.0000 &  0.0000 \\
       yo &  0.0187 &  0.0000 &  0.0000 &  0.0782 &  0.0526 &  0.0000 &  0.0000 \\
\midrule
    means &  0.0189 &  0.0147 &  0.0198 &  0.0268 &  0.0211 &  0.0162 &  0.0126 \\
  medians &  0.0201 &  0.0076 &  0.0120 &  0.0168 &  0.0167 &  0.0120 &  0.0143 \\
\bottomrule
\end{tabular}
\caption{F1 scores for monolingual fine-tuning on the perturbed data for various levels of sparsity with pruning embedding layers.}
\label{tab:mono_perturbed_with_embeddings}
\end{table*}

\begin{table*}
\centering
\scalebox{0.75}{
\begin{tabular}{ll}
\toprule
 & Example English test sentences \\
\midrule 
Original &  Much construction was undertaken during this period , such as the building of Palermo Cathedral . \\
Perturbed & Much construction was undertaken during this period , such as the building of Knott 's Soak City . \\
\midrule
Original &  It is found in Peru . \\
Perturbed & It is found in Carbon Cliff , Illinois . \\
\midrule
Original & Alberto Mancini won in the final 7–5 , 2–6 , 7–6 , 7–5 against Boris Becker. \\
Perturbed & John Jones ( footballer , born 1895 ) won in the final 7–5 , 2–6 , 7–6 , 7–5 against Sultan Ahmad Shah .\\
\midrule
Original &  It flows from Ägerisee through Lake Zug into the Reuss . \\
Perturbed & It flows from New Orleans through Humboldt County , Nevada into the Crow Agency , Montana .\\
\midrule
Original &  The album 's lead single `` Better Believe It '' featuring Young Jeezy and Webbie , was released on July 14 , 2009 . \\
Perturbed & The album 's lead single `` Better Believe It '' featuring W. S. Merwin and Empress Maria Theresa , was released on July 14 , 2009 .\\
\bottomrule
\end{tabular}}
\caption{Example of test sentences for English language using the \textit{entity mention replacement} \cite{dai-adel-2020-analysis} technique where an entity is randomly swapped with another entity of the same type.}
\label{tab:sample_examples_english}
\end{table*}

\begin{table*}
\centering
\scalebox{0.9}{
\begin{tabular}{ll}
\toprule
 & Example Yoruba test sentences \\
\midrule 
 Original & Egb{\'e} Ol{\'o}{\d s}{\`e}l{\'u}ar{\'a}{\'i}l{\'u} {\'a}won Ar{\'a}{\`a}l{\`u} ( N{\`a}ijiri{\'a}  )\\
 Perturbed & Il{\'e}-{\`i}gbim A{\'o}fin On{\'i}b{\'i}nib{\'i} il N{\`a}{\`i}j{\'i}r{\'i}{\`a} ) \\
 
\midrule
Original & Agb{\`e}gb{\`e} {\`I}joba Agb{\`e}gb{\`e} {\`I}j{\d o}ba {\`I}b{\'i}l{\`e} \\
Perturbed & Agb{\`e}gb{\`e} {\`I}joba {\`I}b{\'i}l{\`e} G{\'u}{\'u}s{\`u}-{\`I}w{\`o}r{\`u}n {\`E}k{\`i}t{\`i} Wudil\\
\midrule
Original & {\`A}gb{\'a}j{\d o} {\`a}won Or{\'i}l{\d e}-{\`e}d{\`e} A{\d s}{\`o}kan \\
Perturbed & {\`A}k{\'o}j{\d o} {\`a}won ol{\' o}r{\'i} {\`i}joba il Burkina Faso A{\'o}ka .\\
\midrule

Original & Asia il{\` e} Tufalu .\\
Perturbed & Abdusalami Abubakar Tufalu \\
\midrule
Original & ' '' ' '' j{\' e} F{\' a}r{\' a}{\` o} ni {\' E}g{\' i}pt{\` i} Ay{\' e}ij{\' o}un .\\
Perturbed & ' '' ' '' j Yousaf Raza Gillani ni N{\`a}{\`i}j{\'i}r{\'i}{\`a} .\\
\bottomrule
\end{tabular}}
\caption{Example of test sentences for Yoruba language using the \textit{entity mention replacement} \cite{dai-adel-2020-analysis} technique where an entity is randomly swapped with another entity of the same type.}
\label{tab:sample_examples_yoruba}
\end{table*}

\end{document}